\documentclass{article}

\PassOptionsToPackage{numbers, compress}{natbib}

     \usepackage[preprint]{gpsmdms_2022}

\usepackage[utf8]{inputenc} 
\usepackage[T1]{fontenc}    
\usepackage{hyperref}       
\usepackage{url}            
\usepackage{booktabs}       
\usepackage{amsfonts}       
\usepackage{nicefrac}       
\usepackage{microtype}      
\usepackage{xcolor}         

\usepackage{bbold}
\usepackage{bm}
\usepackage{amsmath}
\usepackage{amssymb}
\usepackage{graphicx}
\usepackage{placeins}
\usepackage{algorithmic, algorithm2e}
\usepackage{subcaption}
\usepackage{natbib}
\usepackage{placeins}
\usepackage{paralist}

\newcommand{\indicator}{\mathbb{1}}
\DeclareMathOperator*{\probit}{probit}

\newcommand{\identity}{\bm{I}}

\newcommand{\fscore}{F\textsubscript{1}}

\DeclareMathOperator*{\argmax}{arg\,max}

\newcommand{\SKIP}[1]{}

\title{An Active Learning Reliability Method for Systems with Partially Defined Performance Functions}

\author{%
  Jonathan Sadeghi \\
  Five AI Ltd. \\
  \texttt{jonathan.sadeghi@five.ai} \\
   \And
   Romain Mueller \\
  Five AI Ltd. \\
   \And
  John Redford \\
  Five AI Ltd. \\
}

\begin{document}

\maketitle

\begin{abstract}
  In engineering design, one often wishes to calculate the probability that the performance of a system is satisfactory under uncertainty.
  State of the art algorithms exist to solve this problem using active learning with Gaussian process models.
  However, these algorithms cannot be applied to problems which often occur in the autonomous vehicle domain where the performance of a system may be undefined under certain circumstances.
  To solve this problem, we introduce a hierarchical model for the system performance, where undefined performance is classified before the performance is regressed.
  This enables active learning Gaussian process methods to be applied to problems where the performance of the system is sometimes undefined, and we demonstrate the effectiveness of our approach by testing our methodology on synthetic numerical examples for the autonomous driving domain.
  The code to generate these experiments is available as an open source repository: \url{https://github.com/fiveai/hGP_experiments/}
\end{abstract}

\section{Introduction}
\label{sec:introduction}

Estimating the probability that the performance of a system is satisfactory under uncertain or variable operating circumstances is an important step towards deploying such systems safely in the real world.
This is especially important in safety critical application such as autonomous driving, where finding rare but catastrophic failures has been identified as a important challenge~\cite{koopman2016challenges}.
Powerful active learning approaches based on Gaussian processes (GP) have been proposed as a solution to this problem~\cite{echard_ak-mcs_2011, yun2019ak, inatsu2021active, iwazaki2021bayesian, moustapha2021generalized, DANG2022108621} and achieve state of the art performance (we describe related work in detail in Section~\ref{sec:related}).
However, such approaches cannot be applied to problems where the performance of the system may be \emph{undefined} under certain specific circumstances, a situation which often occurs in the autonomous vehicle domain~\cite{koopman2016challenges, mahmud, nalic, koopman2018toward}.
For example, consider the case of an autonomous vehicle waiting to join a road at a T-junction, where other cars are travelling on the road at constant velocity.
Assuming that the autonomous vehicle behaves deterministically with respect to variables like the initial starting positions and velocities of the vehicles, one could specify a measure of safe system performance, for example the distance of closest approach between the autonomous vehicle and the other vehicles as a function of these variables.
However, in some cases the closest distance function could be undefined because the autonomous vehicle never joined the road at all, for example if the autonomous vehicle's planner deems the situation to be unsafe.
Na\"ively masking away regions where the performance of the system is undefined would introduce discontinuities and leads to poor performance since Gaussian processes with stationary kernels are not well suited to the regression of discontinuous targets \cite{moustapha2019two}.
In this work, we extend these methods from first principles to the case where the performance function can be undefined by using a hierarchical model (termed hGP) for the system performance, where undefined performance is classified before the performance is regressed.
  
We consider a system whose performance is described by a function $g : \mathcal{X} \mapsto \mathbb{R} \cup \{\textrm{NaN}\}$, where $\bm x \in \mathcal{X} \subseteq \mathbb{R}^k$ are random environmental variables affecting the system, $g(\bm x)<0$ denotes an undesirable event (a failure), and $g(\bm x) = \textrm{NaN}$ is an event of unspecified performance.
An undefined value does not indicate that an undesirable event has occurred for a particular $\bm x$, and therefore we wish to classify these $\bm x$ differently to the failure events.
The rate of failures is quantified using the probabilistic threshold robustness (PTR) of the system, which we define as
\begin{equation}
   p_f = \int_{\mathcal{X}} \indicator\left[g(\bm x) < 0 \cap  g(\bm x) \ne \textrm{NaN} \right] p(\bm x) d \bm x, \label{eqn:modified_ptr}
\end{equation}
where $\indicator$ is the indicator function and $p(\bm x)$ is the probability density (mass) function of $\bm x$ \cite{beyer2007robust}.
Eq.~\eqref{eqn:modified_ptr} represents the probability that the system is in the failure state, while disregarding the `uninteresting' cases where the performance is undefined.
Note that we are not attempting to model the distribution of environment variables and treat $p(\bm x)$ as given.
Estimating $p_f$ in Eq.~\eqref{eqn:modified_ptr} using a vanilla Monte Carlo simulation can be computationally expensive since identifying a failure rate lower than $\epsilon$ will typically require at least $1/\epsilon$ tests \cite{uesato2018rigorous}.
In order to reduce the required number of samples, we use the Adaptive Kriging Monte Carlo Simulation (AK-MCS) algorithm, a simple active learning technique based on Gaussian processes which was shown to provide an extremely efficient evaluation of the PTR measure for previously studied problems~\cite{echard_ak-mcs_2011}, and extend it to partially undefined performance functions.
We perform experiments comparing our proposed methodology to several na\"ive baselines on problems for which the results are known analytically.
We find that our approach produces a more accurate estimation of $p_f$ and also that the surrogate model is a more accurate representation of the true performance function.
\section{Related Work}
\label{sec:related}

The PTR measure has recently become of interest in the robust optimisation literature \cite{beyer2007robust}, for example \citet{inatsu2021active} show how system designs can be adjusted to optimise the measure.
The measure has a much longer history in reliability engineering \cite{melchers2018structural}.
\citet{moustapha2021generalized} and \citet{teixeira2021adaptive} provide reviews of active learning methods for estimating this measure.
The Adaptive Kriging Monte Carlo methodology is perhaps the most well known of these methods, and achieves close to state of the art results \cite{echard_ak-mcs_2011, yun2019ak, moustapha2021generalized}.
Efficient methods of estimating the PTR measure also exist in reinforcement learning \cite{uesato2018rigorous}.
\citet{beglerovic2017testing} use a Bayesian optimisation approach to identify failure cases for an autonomous vehicle but do not exhaustively search for all $\bm x$ such that $f(\bm x) < 0$ and also do not calculate the PTR measure, $p_f$.
Similarly, \citet{urtasun} use a realistic LiDAR simulator to modify real-world LiDAR data which can then be used to test end-to-end autonomous driving systems while searching for adversarial traffic scenarios with Bayesian Optimisation.
A related problem in the autonomous vehicle space is finding the most likely $\bm x$, i.e. with largest $p( \bm x)$, leading to $f(\bm x) < 0$ \cite{koren2019adaptive}.
This is closely related to first order methods for estimating the PTR measure \cite{melchers2018structural}.

Although there exists literature related to Gaussian Process modelling for discontinuous targets \cite{moustapha2019two}, there is little literature on active learning specifically for the PTR measure for discontinuous targets.

\section{Approach}
\label{sec:methodology}

We modify the AK-MCS active learning algorithm by using a different Gaussian Process and acquisition function to \citet{echard_ak-mcs_2011}.
We use a hierarchical Gaussian process model for the rule function, consisting of separate regression and classification Gaussian processes, and a modified acquisition function which minimises the catastrophic event classification error to yield an optimal surrogate model of the rule function.
Otherwise our proposed algorithm proceeds in the same way as the AK-MCS algorithm, i.e. an initial training set is chosen to train a Gaussian process, and then subsequent evaluations of the performance function, $g$, are chosen iteratively by maximising a function of the Gaussian process known as the acquisition function, which are then used to retrain the Gaussian process.
The algorithm terminates when the coefficient of variation (CoV) of the failure probability computed using the Gaussian process is below some threshold, and the predicted misclassification probability is also below some threshold.
This algorithm is shown in Algorithm~\ref{alg:1}.

Let $y_*$ be the predicted performance for the test input $\bm x_* \in \mathcal{X}$ where $y \in \mathbb{R} \cup \{\textrm{NaN}\}$, and let the dataset of training examples $\mathcal{D} = \{ (\bm x_i, y_i) | i= 1,...,n\}$.
We model the predictive distribution $p(y_* | \bm x_*, \mathcal{D})$ hierarchically as
\begin{equation}
       p(y_* | \bm x_*, \mathcal{D}) =
    \begin{cases}
    p(y_* | \bm x_*, \mathcal{D}, y_* \ne \textrm{NaN}) p(y_* \ne \textrm{NaN} | \bm x_*, \mathcal{D}) & \text{if $y_* \ne \textrm{NaN}$}, \\
    p(y_* = \textrm{NaN} | \bm x_*, \mathcal{D}) & \text{if $y_* = \textrm{NaN}$},
    \end{cases} 
    \label{eq:hierarchical_model}
\end{equation}
where $ p(y_* | \bm x_*, \mathcal{D}, y_* \ne \textrm{NaN})$ is the predicted regression distribution for $y_*$ at the test input $\bm x_*$ given that $y_*$ is defined, and $p(y_* = \textrm{NaN} | \bm x_*, \mathcal{D})$ is the predicted classification probability that $y_*$ is undefined for $\bm x_*$.
We model these distributions with separate GPs; for $p(y_* |  \bm x_*, \mathcal{D},  y_* \ne \textrm{NaN})$ GP regression is used, and for $p(y_* = \textrm{NaN} | \bm x_*, \mathcal{D})$ GP classification is used.
The conditional prediction of the failure event can easily be calculated as
$
     p(y_* < 0, y_* \ne \textrm{NaN}| \bm x_*, \mathcal{D}) 
$,
which can be used to define an acquisition function, $p_\text{misclassification}(\bm x)$, based on probability of misclassification of $y_* < 0 \cap  y_* \ne \textrm{NaN}$, as in \citet{echard_ak-mcs_2011}.
We give more details about our modelling approach in Appendix~\ref{sec:appendix_method}.
Finally, our hierarchical model Eq.~\eqref{eq:hierarchical_model} can be used to compute the failure probability as
\begin{equation}
   p_f \approx \int_\mathcal{X} {\indicator[p(y_* < 0, y_* \ne \textrm{NaN}| \bm x_*, \mathcal{D}) > 0.5]}  p(\bm x_*) d \bm x_* \label{eqn:gp_pf}.
\end{equation}
The termination criteria for the AK-MCS algorithm will determine the error in the failure probability computed using the Gaussian process in Eq.~\eqref{eqn:gp_pf}, in addition to bounding the error of the hierarchical Gaussian process model.
This ensures that the model is sufficiently accurate to be used by engineers to make predictions about the behaviour of the system.

Note that in this paper we only consider systems with a single performance function, however \citet{yun2019ak} demonstrate how acquisition functions for multiple performance functions can be combined when one is interested in a combined PTR measure for the performance functions.
This can be applied to our hierarchical model directly.
Finally, we note that our modifications to the AK-MCS algorithm are fairly general and only involve changing the performance function model and acquisition function, and therefore these changes could possibly also be used with different active learning algorithms.
We do not explore these possible applications in this paper and instead leave this as a topic for future research.

\RestyleAlgo{ruled}
\begin{algorithm}
    \caption{Hierarchical Gaussian Process PTR Active Learning Method}
    \label{alg:1}
    \begin{algorithmic}
        \REQUIRE GP prior $\mathcal{GP}(0, k)$, termination threshold $\eta$, model $g(\bm x)$
        \STATE Define proposal set $\mathcal{S}$: sample $n_{mc}$ points from $p(\bm x_*)$.
        \STATE Define initial design of experiments: sample $n_E$ points uniformly from $\mathcal{S}$ and evaluate with model $g(\bm x)$ to define $\hat{\mathcal{S}} = \{ (\bm x_i, y_i) | i= 1,...,n_E\}$
        \WHILE{$CoV > 0.1$}
            \WHILE{$\max_{\bm x} p_\text{misclassification}(\bm x) > \eta$}
            \STATE Train GP on $\hat{\mathcal{S}}$
            \STATE Compute $\mu(\bm{x}),~\sigma(\bm{x}),~ p_\text{nan}(\bm x)$ from hierarchical GP for all $\bm{x} \in \mathcal{S}$.
            \STATE Choose most likely misclassified $\bm x$: $\bm{x}_* = \argmax_{\bm x \in \mathcal{S}} p_\text{misclassification}(\bm x)$ 
            \STATE Observe $y_* = g(\bm{x}_*)$ and Add $(\bm{x}_*, y_*)$ to $\hat{\mathcal{S}}$
        \ENDWHILE
        \STATE Estimate $p_f$ using Monte Carlo simulation with Gaussian Process on $\mathcal{S}$ using Eq.~\eqref{eqn:gp_pf}
        \STATE Calculate $CoV = \sqrt{\frac{(1-p_f)}{p_f |\mathcal{S}|}}$
        \STATE Sample $n_{mc}$ points from $p(\bm x_*)$ and evaluate with model $g(\bm x)$ to add to $\mathcal{S}$
        \ENDWHILE
        \ENSURE Fitted hierarchical GP and $p_f$ computed using Eq.~\eqref{eqn:gp_pf}.
    \end{algorithmic}
\end{algorithm}

\section{Experiments}
\label{sec:experiments}

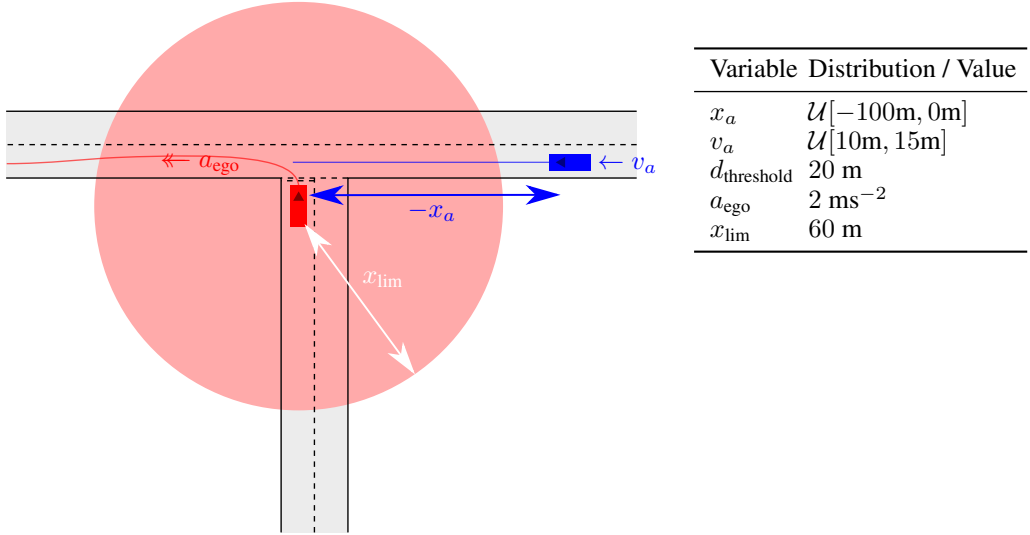
\begin{figure}
    \centering
    \def\svgwidth{0.6\textwidth}
\newcommand{\doubleleftarrow}{%
  \leftarrow\mathrel{\mspace{-15mu}}\leftarrow
}

\begingroup%
  \makeatletter%
  \providecommand\color[2][]{%
    \errmessage{(Inkscape) Color is used for the text in Inkscape, but the package 'color.sty' is not loaded}%
    \renewcommand\color[2][]{}%
  }%
  \providecommand\transparent[1]{%
    \errmessage{(Inkscape) Transparency is used (non-zero) for the text in Inkscape, but the package 'transparent.sty' is not loaded}%
    \renewcommand\transparent[1]{}%
  }%
  \providecommand\rotatebox[2]{#2}%
  \newcommand*\fsize{\dimexpr\f@size pt\relax}%
  \newcommand*\lineheight[1]{\fontsize{\fsize}{#1\fsize}\selectfont}%
  \ifx\svgwidth\undefined%
    \setlength{\unitlength}{345.78187477bp}%
    \ifx\svgscale\undefined%
      \relax%
    \else%
      \setlength{\unitlength}{\unitlength * \real{\svgscale}}%
    \fi%
  \else%
    \setlength{\unitlength}{\svgwidth}%
  \fi%
  \global\let\svgwidth\undefined%
  \global\let\svgscale\undefined%
  \makeatother%
  \begin{picture}(1,0.84213778)%
    \lineheight{1}%
    \setlength\tabcolsep{0pt}%
    \put(0,0){\includegraphics[width=\unitlength,page=1]{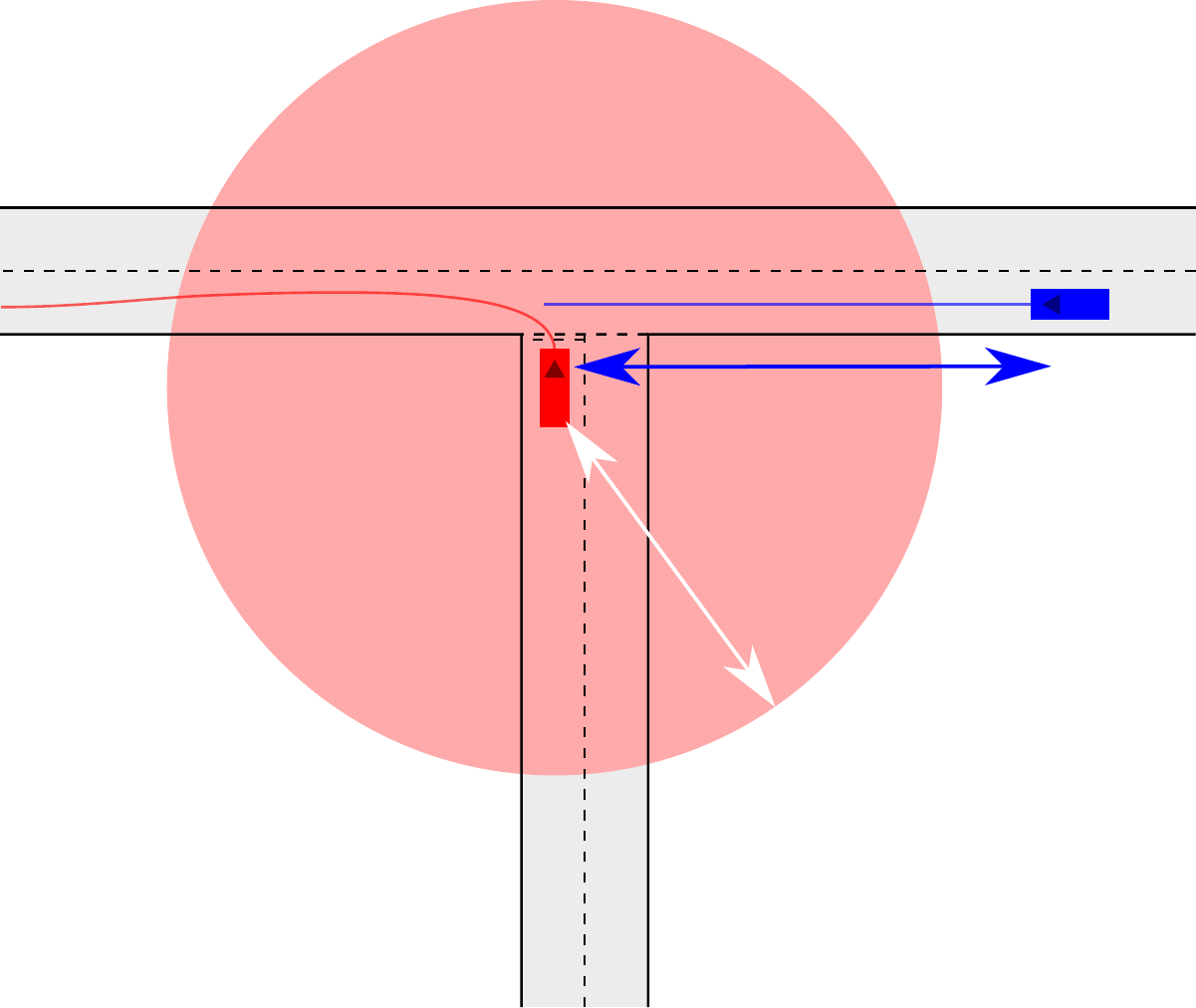}}%
    \put(0.93842594,0.57677331){\color[rgb]{0,0,1}\makebox(0,0)[lt]{\lineheight{1.25}\smash{\begin{tabular}[t]{l}$\leftarrow  v_a$\end{tabular}}}}%
    \put(0.63713205,0.50059742){\color[rgb]{0,0,1}\makebox(0,0)[lt]{\lineheight{1.25}\smash{\begin{tabular}[t]{l}$-x_a$\end{tabular}}}}%
    \put(0.56519222,0.39570286){\color[rgb]{1,1,1}\makebox(0,0)[lt]{\lineheight{1.25}\smash{\begin{tabular}[t]{l}$x_\text{lim}$\end{tabular}}}}%
    \put(0.24492494,0.58){\color[rgb]{1,0,0}\makebox(0,0)[lt]{\lineheight{1.25}\smash{\begin{tabular}[t]{l}$\doubleleftarrow a_\text{ego}$\end{tabular}}}}%
  \end{picture}%
\endgroup%

    \hspace{0.5cm}
    \raisebox{5cm}{
    \begin{tabular}{l@{\hspace{4pt}}l@{\hspace{4pt}}}
    \toprule
         Variable & Distribution / Value \\
    \midrule
         $x_a$ & $\mathcal{U}[-100\text{m}, 0\text{m}]$ \\
         $v_a$ & $\mathcal{U}[10 \text{m}, 15\text{m}]$ \\
         $d_\text{threshold}$ & $20$ m \\
         $a_\text{ego}$ & $2$ ms$^{-2}$ \\
         $x_\text{lim}$ & $60$ m \\
    \bottomrule
    \end{tabular}
    }
    \caption{Depiction of our T-junction experiment.
    Left: Ego is shown in red merging into a road in a left hand traffic system where the adversarial car is shown in blue. The red circle represents the limits of the perception systems of ego. $d_\text{threshold}$, the smallest safe distance between ego and the adversarial car, is not shown.
    Right: Random variables and parameters.}
    \label{fig:tjunctionexperiment}
\end{figure}

\begin{figure}
    \centering
\begin{subfigure}[t]{0.49\textwidth}
    \centering
    \includegraphics[width=\textwidth]{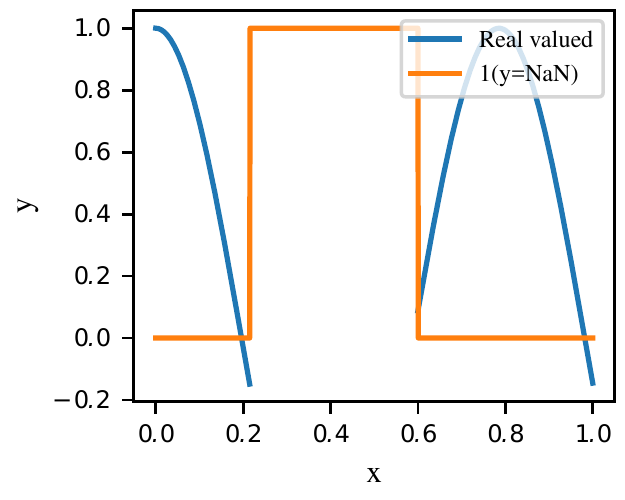}
    \caption{System performance for toy function experiment (Eq.~\eqref{eqn:g}).}
    \label{fig:gt}
\end{subfigure}
\hfill
\begin{subfigure}[t]{0.49\textwidth}
    \centering
    \includegraphics[width=\textwidth]{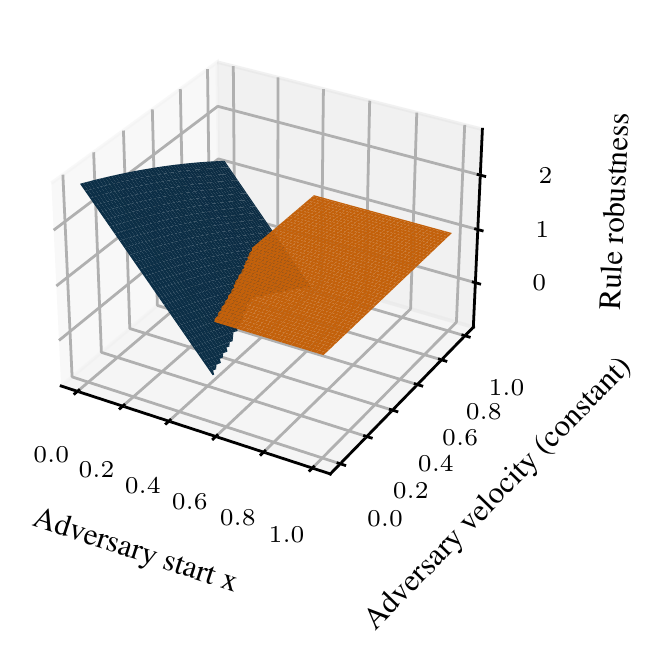}
    \caption{System performance for T-junction merging experiment (Eq.~\eqref{eq:t_performance} with input features normalised). The blue surface shows the real valued performance and the orange surface shows the undefined performance masked with $\alpha=1.0$.}
    \label{fig:gt2}
\end{subfigure}
\caption{Plots of ground truth performance functions.}
\end{figure}

\paragraph{Benchmark tasks:} We evaluate our methodology on two benchmark problems where the system performance is partially undefined and for which $p_f$ can be calculated analytically:
\begin{compactitem}
    \item \emph{Toy function}
    The system performance (plotted in Fig.~\ref{fig:gt}) is given by
\begin{equation}
    g(\bm x) =  
    \begin{cases}
        \text{NaN} & \text{if $0.215 < \bm x < 0.6$}, \\
        \cos(8 \bm x) & \text{otherwise},
    \end{cases} \label{eqn:g}
\end{equation}
where the uncertain variable $x$ is distributed with $p(\bm x) = \mathcal{U}[0,1]$.
    \item \emph{Autonomous driving (AD)} A mathematical model of an autonomous vehicle joining a main road at a T-junction, as depicted in Fig.~\ref{fig:tjunctionexperiment}.
    The thresholded minimum lateral distance between the autonomous vehicle and an approaching vehicle is given by the performance function (plotted in Fig.~\ref{fig:gt2})
    \begin{equation}
        g(x_a, v_a) =  
    \begin{cases}
        \text{NaN} & \text{if $d_\text{min}(x_a, v_a) < d_\text{threshold}$ and $|x_a| < x_\text{lim}$}, \\
        d_\text{min}(x_a, v_a) - d_\text{threshold} & \text{otherwise},
    \end{cases} \label{eq:t_performance}
\end{equation}
where we have defined the closest approach distance between the vehicles as
$
    d_\text{min}(x_a, v_a) =
    \max\left(
        -\left(x_a + v_a^2/(2 a_\text{ego})\right),
        0
        \right) 
$    
where $v_a$,  $x_a$, $x_\text{lim}$, $d_\text{threshold}$, $a_\text{ego}$ and $v_a$ are physical variables parameterised by distributions shown in Fig.~\ref{fig:tjunctionexperiment}.
A full derivation of the performance function and explanation of physical variables is provided in Appendix~\ref{sec:motivation_model}.
Undefined values here represent scenarios when the autonomous vehicle decides not to join the main road, and so the lateral distance along the road is undefined.
\end{compactitem}

\paragraph{Baselines:}
The hierarchical GP (hGP) will be compared with the following baseline methodologies:
\begin{compactitem}
    \item \emph{Masked GP}: AK-MCS \cite{echard_ak-mcs_2011} with a regression GP where NaN values are masked with positive constant $\alpha>0$ , i.e.
\begin{equation}
      \tilde{g} = \begin{cases} 
      g(\bm x)  & \text{if $g(\bm x) \ne \textrm{NaN}$}, \\  
      \alpha  & \text{if $g(\bm x) = \textrm{NaN}$}. \end{cases}
\end{equation}
    \item \emph{Active GP Classification} (GPC): similarly to \citet{kapoor2007active}, we apply a GP classifier to classify the event $y_* < 0 \cap y_* \ne \textrm{NaN}$ and use this to replace the Kriging model in the AK-MCS active learning loop.
\end{compactitem}
Hyperparameters for each algorithm are shown in Appendix~\ref{appendix:hyperparams}.
We repeat the experiments with different values of $\alpha$ ($\alpha = 0.1$, $\alpha = 0.5$, $\alpha = 1.0$) for the Masked GP in Appendix~\ref{sec:alpha}, where no significant differences are observed for different values of $\alpha$ (results for $\alpha=1$ are shown in this section).

\paragraph{Metrics:}
We use the following metrics, some of which were used in the original AK-MCS paper \cite{echard_ak-mcs_2011}, and some of which we introduce specifically to measure aspects of performance related to our problem.
\begin{itemize}
    \item Maximum predicted misclassification ($\max_{\bm x \in \mathcal{S}} p_\text{misclassification}(\bm x)$) bounds the "risk" suffered over $p(\bm x)$, and hence can be used to measure the convergence of the \emph{internal} state of the algorithm, i.e. it does not require ground truth data. Used in \citet{echard_ak-mcs_2011}.
    \item Failure probability ($p_f$) can be used to measure if the identified failure region has the correct size, by comparing to the $p_f$ computed in some other way as ground truth (i.e. analytically). Used in \citet{echard_ak-mcs_2011}.
    \item Coefficient of Variation (CoV) is used to assess the algorithms internal uncertainty in the estimated $p_f$, and can be calculated as described in Algorithm~\ref{alg:1}. Used in \citet{echard_ak-mcs_2011}.
    \item \fscore score can be used to check if the identified failure region is correctly located in the space of $x$. We introduce this metric because in reliability problems the failure probability is usually low and hence the class distribution is imbalanced, and \fscore score is known to perform well in such cases. The \fscore score is defined as $F_1 = \frac{\text{precision} \times \text{recall}}{\text{precision} + \text{recall}}$ where precision and recall are measured with a test set of $10^5$ points, treating failure as the positive class and the safe region as the negative class. Used in \citet{iwazaki2021bayesian}.
    \item Average precision penalises an incorrectly located failure region in a similar way to \fscore score, however it has the additional advantage that the ranking of the predicted class scores is tested. Average precision is insensitive to correctly ranked but miscalibrated scores. It is also measured with a test set of $10^5$ points.
\end{itemize}

\begin{figure}
     \centering
     \includegraphics[width=.98\textwidth]{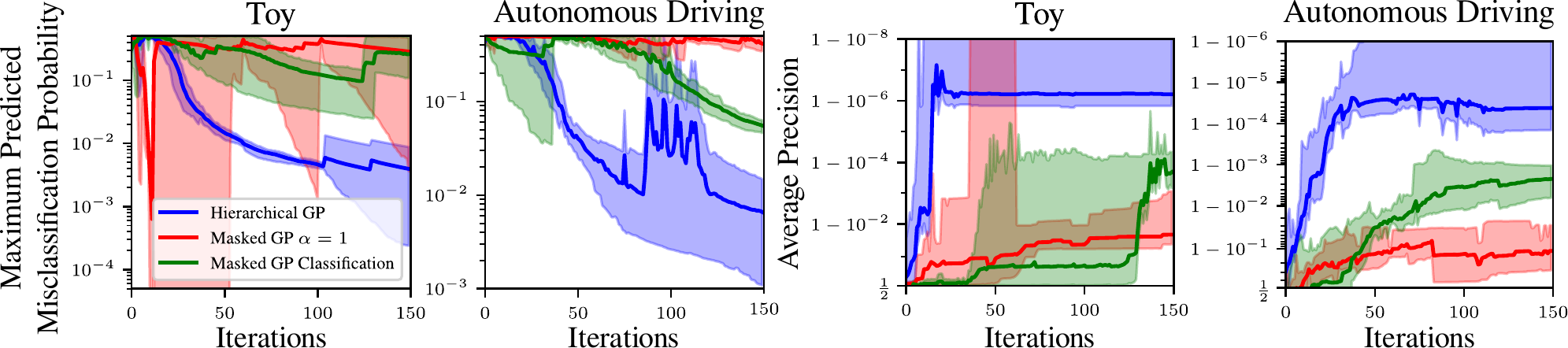}
             \caption{Convergence of average precision and maximum predicted misclassification probability for fitted GPs for the toy function and AD experiments. The shaded area represents the minimum and maximum of 5 repeats, and the dark line represents the mean.).
        }
        \label{fig:convergence_squashed}
\end{figure}

\begin{figure}
    \centering
      \includegraphics[width=.97\textwidth]{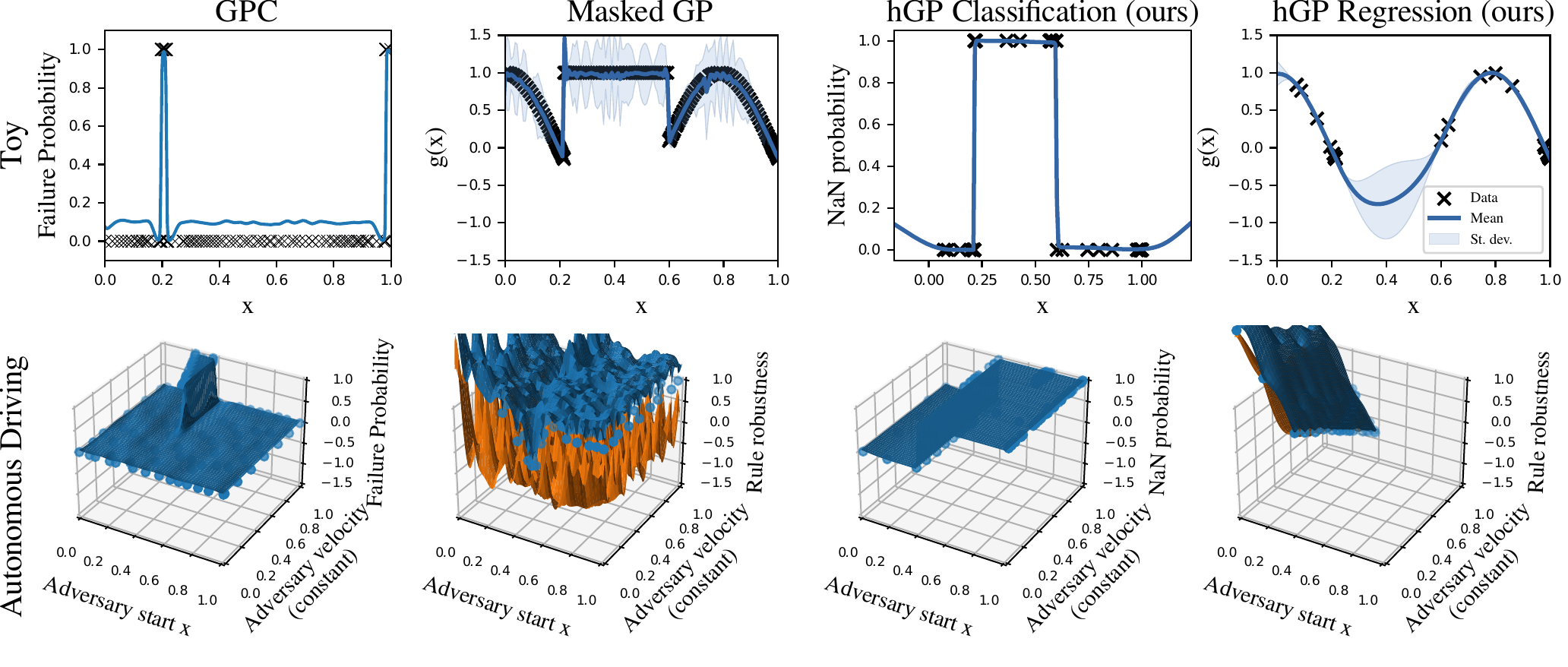}
    \caption{Plotted models for the toy function and AD experiments.
    One standard deviation prediction bounds are shown for the AD regression GPs as orange and blue surfaces and for the toy function regression GPs in shaded blue.
    For classification GPs only the predicted probability is shown.
        Training data is shown as scatter points.
    }
    \label{fig:squashed_models}
\end{figure}

\paragraph{Model Accuracy:}
Firstly we run each algorithm for 150 iterations, i.e. neglect the termination criteria by setting $\eta=0$, in order to study the accuracy of the models independently of the termination criteria.
Fig.~\ref{fig:convergence_squashed} shows the convergence of the failure probability and maximum predicted misclassification probability for the models for the Toy and Autonomous experiments.
In addition, in Appendix~\ref{sec:additional_figures} the convergence of the failure probability ($p_f$) and \fscore score for the models is shown.
We observe that the convergence for the hGP is far faster than for the other models.
The convergence of $p_f$ and predicted misclassification probability for the masked GP is erratic, which is indicative of the misspecification of the model.
Visualising the fitted models in Fig.~\ref{fig:squashed_models}, reveals that the low \fscore score and average precision are caused by the length scale for the masked GP becoming extremely short due to the discontinuity in the masked performance function, resulting in a large predicted variance and hence erroneous class scores.
hGP outperforming GPC is unsurprising, as the hGP utilises more information about the magnitude of the performance function in order to make a more educated selection of the next point to query.

Overall, although all models eventually accurately estimate $p_f$, the hGP clearly provides a more accurate classification of the failure region, more stable training, and class probabilities which better represent the state of knowledge given the available data.

\paragraph{Termination Criteria:}
We analyse the ability of the models to terminate the active learning loop after an appropriate number of iterations by repeating the experiments in the previous section with the termination criteria enabled.
Table~\ref{tab:pf_terminate_mini} shows the number of samples, the $p_f$, and \fscore score when the algorithm terminates.
Appendix~\ref{sec:additional_figures} shows additional properties of the models upon termination.
The hGP AK-MCS terminates with an accurate model in both experiments.
For both experiments we see that GPC eventually learns an accurate model but does not terminate, because the maximum predicted misclassification probability remains high.
In the toy function experiment the masked GP terminates far too early while the \fscore score is very low compared to hGP.
The masked AK-MCS does not terminate for the AD experiment; the short length scale in the masked GP means that the predicted variance is large and a large amount of iterations will be required to significantly reduce this.
For the AD experiment all models obtain the correct value of $p_f$ within estimated error, however the \fscore score is much lower for the masked GP, suggesting the identified failure region is incorrectly located.

It is clear that due to the ability of the hGP to predict appropriate classification probabilities it is the only model which can reliably terminate when the model is sufficiently accurate.

\begin{table}
    \centering
    
    \caption{Comparison of $p_f$, \fscore score (\fscore), and number of evaluations (N. Eval.) for all methodologies. N. Eval. includes the initial design of experiments (12 evaluations) and the number of iterations of the active learning algorithm. Number of iterations was capped at 150 and methods hitting the cap are marked with did not terminate (DNT). 
    Mean and standard deviation for 5 repeats are shown.
         }
         \scalebox{0.95}{
		\begin{tabular}{@{}l@{\hspace{4pt}}l@{\hspace{4pt}}l@{\hspace{4pt}}l@{}}
		\toprule
         Methodology & $p_f$ Avg. (Std. Dev.) & \fscore &  N. Eval. \\
         \midrule
          & \multicolumn{3}{c}{Toy function} \\
          \cmidrule{2-4}
         Analytic & 0.036906 & N/A & N/A \\ 
hGP (ours) & 0.036 (0.0051) & 1 (0.0011) & 56 (3.4) \\
Masked GP 
& 0.019 (0.0031) & 0.66 (0.00013) & 20 (3.2) \\
GPC & 0.037 (0.0026) & 0.99 (0.003) & DNT \\
         \bottomrule
    \end{tabular}
    		\begin{tabular}{@{}l@{\hspace{4pt}}l@{\hspace{4pt}}l@{\hspace{4pt}}l@{}}
		\toprule
         $p_f$ & \fscore &  N. Eval. \\
         \midrule
         \multicolumn{3}{c}{Autonomous Driving} \\
          \cmidrule{1-3}
       0.0382 & N/A & N/A  \\ 
0.038 (0.0019) & 1 (0.0013) & 81 (21) \\
0.037 (0.0041) & 0.92 (0.063) & DNT \\
0.035 (0.0029) & 0.98 (0.0049) & DNT \\
         \bottomrule
    \end{tabular}
    }
    \label{tab:pf_terminate_mini}
\end{table}

\section{Conclusion}
In this paper we presented an approach for active learning based Monte Carlo simulation with hierarchical Gaussian process models of partially defined performance functions, and motivate this problem in an autonomous driving context.
We showed that it is possible to construct a Gaussian process model for such a performance function, and derive an acquisition function based on the probability of misclassification for undesirable system performance.
We provided extensive results on synthetic functions to show that our methodology is superior to na\"ive baselines.

In future work the integration of the hierarchical model into other active learning algorithms could be explored, in addition to testing different termination criteria for the algorithm.
In addition, it would be of interest to study the convergence of the algorithm theoretically, and also to extend the proposed algorithm to simultaneously optimise over possible designs of the system as in \citet{iwazaki2021bayesian}.

\FloatBarrier
\begin{ack}
We gratefully acknowledge John Prater, Jamie McCallion, Tom Westmacott, and Iain Whiteside for valuable discussions at the inception of this project. 
We also wish to thank Majd Hawasly for spending his valuable time to provide comments on the work.
\end{ack}

{\small
\bibliographystyle{unsrtnat}
\bibliography{refs}
}


\appendix

\section{Mathematical Model}
\label{sec:motivation_model}
We consider a vehicle moving in the nearside lane at a constant velocity $v_a$, which has starting longitudinal position $x(t=0) = x_a < 0$ relative to the ego vehicle.
The ego vehicle will perceive the adversarial vehicle if the distance between the two vehicles is less than the limiting distance for the sensor ($x_\text{lim}$), otherwise a false negative detection will occur.
The scenario will be considered safe provided the distance between the vehicles in the same lane is no less than a threshold $d_\text{threshold}$, corresponding to the stopping distance of a typical vehicle.
Ego will attempt to merge into the nearside lane and accelerate at its maximum velocity $a_\text{ego}$ until it reaches $v_a$, however if ego perceives that this action will result in a collision due to the position and velocity of the adversarial vehicle then ego will not merge into the road and hence the longitudinal distance between the vehicles in the lane will be undefined.
This is shown in Fig.~\ref{fig:tjunctionexperiment}.

Modelling each vehicle as a particle moving in a one dimensional space, we can write the rule numerically as
\begin{equation}
            g(x_a, v_a) =  
    \begin{cases}
        \text{NaN} & \text{if $d_\text{min}(x_a, v_a) < d_\text{threshold}$ and $|x_a| < x_\text{lim}$}, \\
        d_\text{min}(x_a, v_a) - d_\text{threshold} & \text{otherwise},
    \end{cases}
\end{equation}
where we have defined the closest approach distance between the vehicles as
\begin{equation}
    d_\text{min}(x_a, v_a) = \min_{t \in [0, \infty]} \left|\frac{1}{2} a_\text{ego} t^2 - (x_a + v_a t)\right| = 
    \max\left(
        -\left(x_a + \frac{v_a^2}{2 a_\text{ego}}\right),
        0
        \right)
        .
\end{equation}
In order to ensure that the scale of the performance function is appropriate for the GP hyperparameters we have chosen, we rescale the performance function by dividing by 20 resulting in performance values of lower magnitude.

\section{Hierarchical model details}
\label{sec:appendix_method}

In this section we provide additional details which describe the model proposed in Section~\ref{sec:methodology} in further detail.
Let $\bm X$ be a matrix containing all the $\bm x$ in $\mathcal{D}$ and $Y$ be a vector containing all the $y$ in $\mathcal{D}$.
We assume that $p(y_* |  \bm x_*, \mathcal{D},  y_* \ne \textrm{NaN})$ follows a Gaussian process, i.e. $p(y_* |  \bm x_*, \mathcal{D},  y_* \ne \textrm{NaN}) = \mathcal{N}(\mu(\bm x_*), \sigma(\bm x_*))$ with
\begin{align*}
    \mu(\bm{x}_*) &= {\bm k}(\bm{x}_*, \bm{X})^{\top}
    ({\bm k}(\bm{X}, \bm{X}) + \sigma_\text{noise}^2 \identity)^{-1} {Y}, \\
    \sigma^{2}(\bm{x_*} ) &= {\bm k}(\bm{x}_*, \bm{x}_*) -
    {\bm k}(\bm{x}_*, \bm{X}) ^{\top}({\bm k}(\bm{X}, \bm{X}) + \sigma_\text{noise}^2 \identity)^{-1} {\bm k}(\bm{X}, \bm{x}_*) ,
\end{align*}
where $k: \mathcal{X} \times \mathcal{X} \rightarrow \mathbb{R}$
is a positive definite kernel with $ 0< k(\bm{x}, \bm{x}) \leq 1$ for all $\bm{x} \in \mathcal{X}$, and $K(\bm X_1, \bm X_2)$ is matrix containing evaluations of the kernel at all points in $\bm X_1$ and $\bm X_2$, and $\sigma_\text{noise}$ is a small positive constant which should be inversely proportional to how deterministic the evaluation of $y$ is.

We assume that  $p(y_* = \textrm{NaN} | \bm x_*, \mathcal{D})$ is given by a classification Gaussian process, i.e. $p(y_* = \textrm{NaN} | \bm x_*, \mathcal{D}) = \int \probit(h) p(h | \bm x_*, \mathcal{D}) dh = p_\text{nan}(\bm x_*)$, where $ p(h | \bm x_*, \mathcal{D})$ is the predictive distribution of a Gaussian process which can be calculated using the expectation propagation method as described in \citet{williams2006gaussian}. 

To calculate the predicted probability of misclassification for our model, recall that we are trying to define the classification boundary between the failure event $y_* < 0 \cap y_* \ne \textrm{NaN}$ and the complementary case.
We classify this event based on $p(y_* < 0,  y_* \ne \textrm{NaN} | \bm x_*, \mathcal{D}) = p(y_* < 0 | \bm x_*, \mathcal{D}, y_* \ne \textrm{NaN}) p(y_* \ne \textrm{NaN} | \bm x_*, \mathcal{D}) > 0.5$.
Therefore we calculate the predicted misclassification probability as
\begin{equation}
    p_\text{misclassification}(x_*) =  
    \begin{cases}
        p(y_* < 0,  y_* \ne \textrm{NaN} | \bm x_*, \mathcal{D}) & \text{if $p(y_* < 0,  y_* \ne \textrm{NaN} | \bm x_*, \mathcal{D}) < 0.5$}, \\
        1 - p(y_* < 0,  y_* \ne \textrm{NaN} | \bm x_*, \mathcal{D})  & \text{otherwise}.
    \end{cases} \label{eqn:acquisition}
\end{equation}
where $p(y_* < 0,  y_* \ne \textrm{NaN} | \bm x_*, \mathcal{D}) = \Phi \left( - {\mu}(\bm x_*)/\sigma(\bm x_*) \right) p(y_* \ne \textrm{NaN} | \bm x_*, \mathcal{D}) $,
where ${\mu}(\bm x_*)$ and ${\sigma}(\bm x_*)$, are the predicted mean and standard deviation of the regression Gaussian process, and $\Phi$ is the standard normal CDF.

We calculate the conditional failure probability for the regression Gaussian process using $
     p(y_* < 0 | \bm x_*, \mathcal{D}, y_* \ne \textrm{NaN}) = \Phi \left( - {\mu}(\bm x_*)/\sigma(\bm x_*) \right)
$, where ${\mu}(\bm x_*)$ and ${\sigma}(\bm x_*)$, are the predicted mean and standard deviation of the regression Gaussian process.
It follows when $p(y_* \ne \textrm{NaN} | \bm x_*, \mathcal{D}) = 1$, Eq.~\eqref{eqn:acquisition} reduces to the form in \citet{echard_ak-mcs_2011}, i.e. $ p_\text{misclassification}(\bm x_*) = \Phi \left( - |\mu(\bm x_*)|/\sigma(\bm x_*) \right)$.

\section{Experimental Hyperparameters}
\label{appendix:hyperparams}
For the classification part of the hierarchical GP we use a Matern52 kernel and fix the variance to $10^5$ as we find that this ensures a quick convergence in practice, and corresponds to a prior belief that the classification GP should model a deterministic function.
For all regression GPs we use a Matern52 kernel with variance and lengthscale determined by optimisation on the training data with some relatively weak constraints: the lengthscale falls in $[0, 0.2]$ and variance falls in $[0.5, 1]$.
For GPC we use a Matern52 kernel with variance $100$ and length scale constrained within $[0, 0.2]$.
For all GPs we set the likelihood variance to a small positive number ($0.005^2$), representing a belief that the system performance is deterministic.
For AK-MCS we add $n_{mc} = 5 \times 10 ^ 3$ proposal samples when the maximum misclassification is below $\eta = 0.02$ and terminate the algorithm when the coefficient of variation is below 0.1, which are similar criteria to those used in \citet{echard_ak-mcs_2011}.
In all experiments, the initial design of experiments is 12 samples.
The Gaussian Process models were created using GPy \cite{gpy2014}, and Emukit was used for the active learning algorithms \cite{emukit2019}.

\section{Additional Experimental Results}

\subsection{Additional Figures}
\label{sec:additional_figures}
\begin{figure}
     \centering
     \begin{subfigure}[b]{0.49\textwidth}
         \centering
         \includegraphics[width=\textwidth]{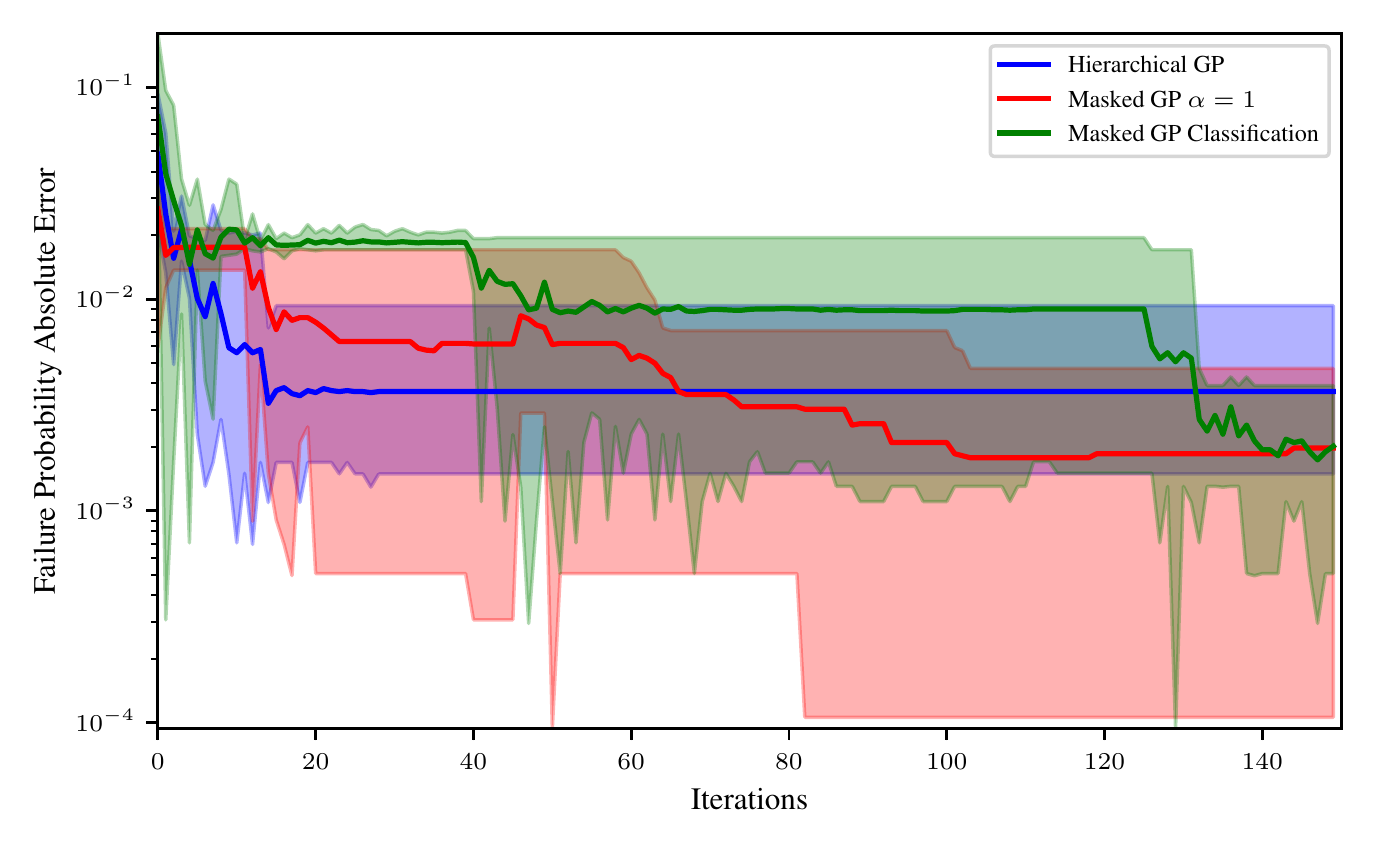}
     \end{subfigure}
     \hfill
     \begin{subfigure}[b]{0.49\textwidth}
         \centering
         \includegraphics[width=\textwidth]{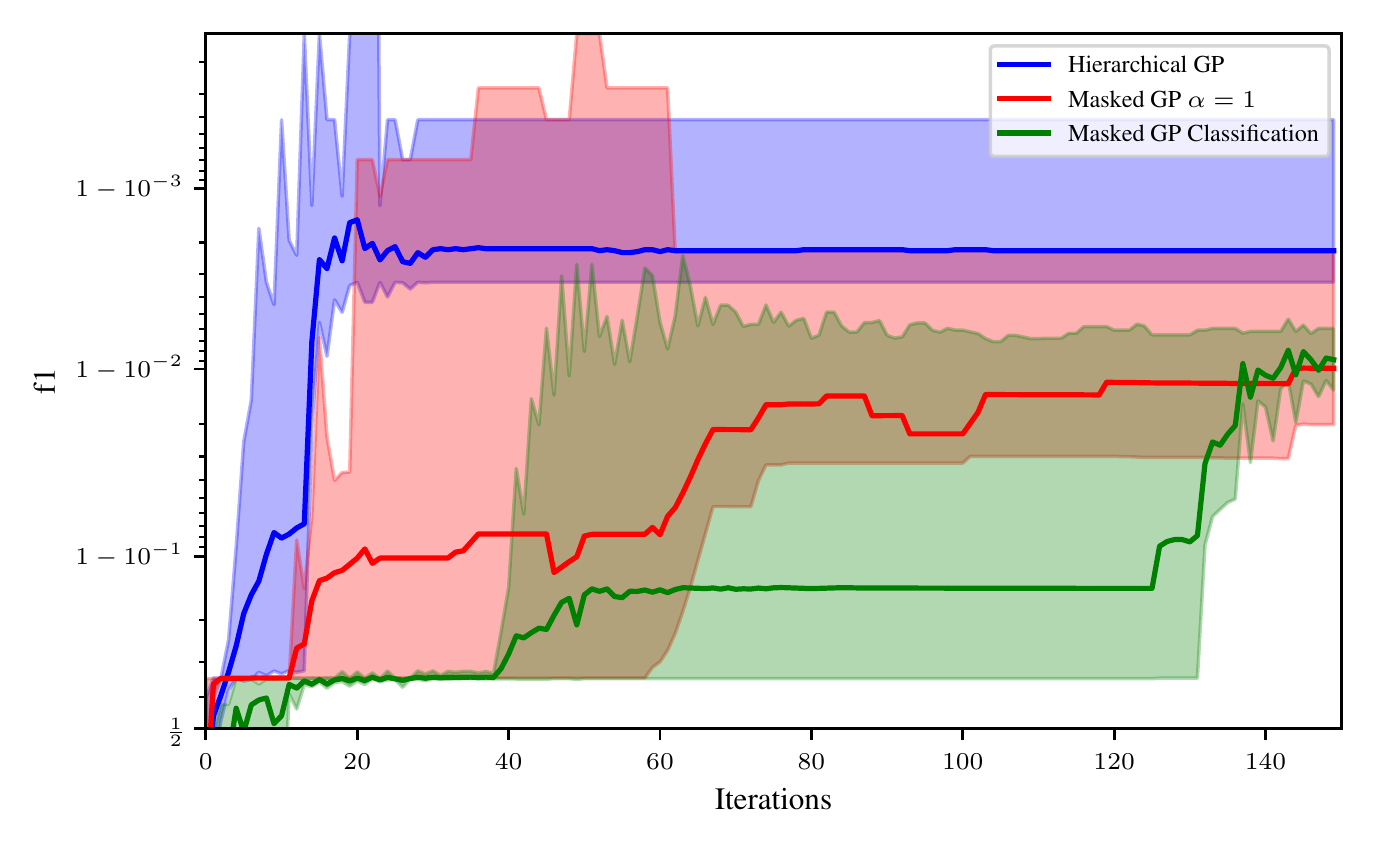}
     \end{subfigure}\\
          \begin{subfigure}[b]{0.49\textwidth}
         \centering
         \includegraphics[width=\textwidth]{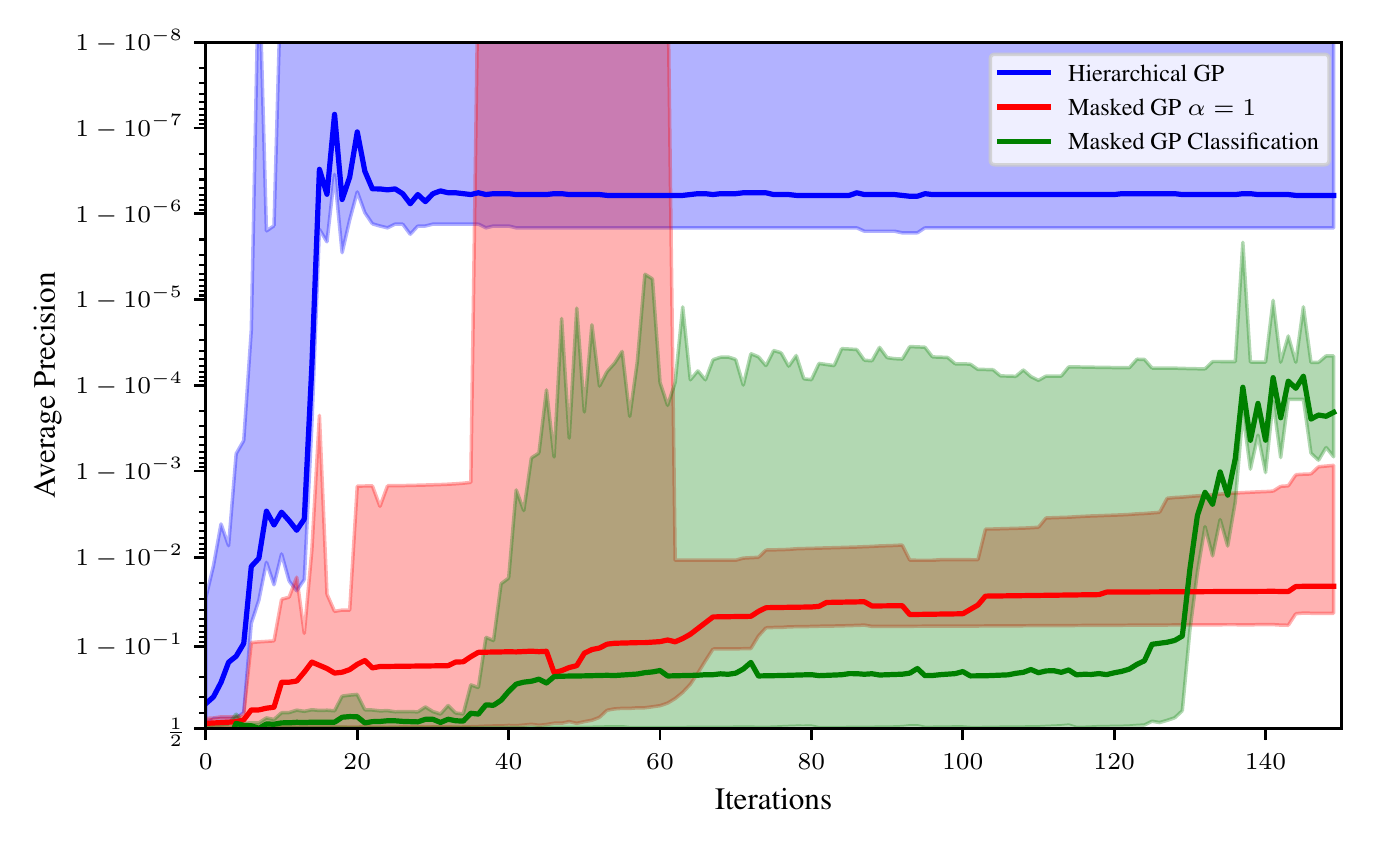}
   
     \end{subfigure}
          \hfill
     \begin{subfigure}[b]{0.49\textwidth}
         \centering
         \includegraphics[width=\textwidth]{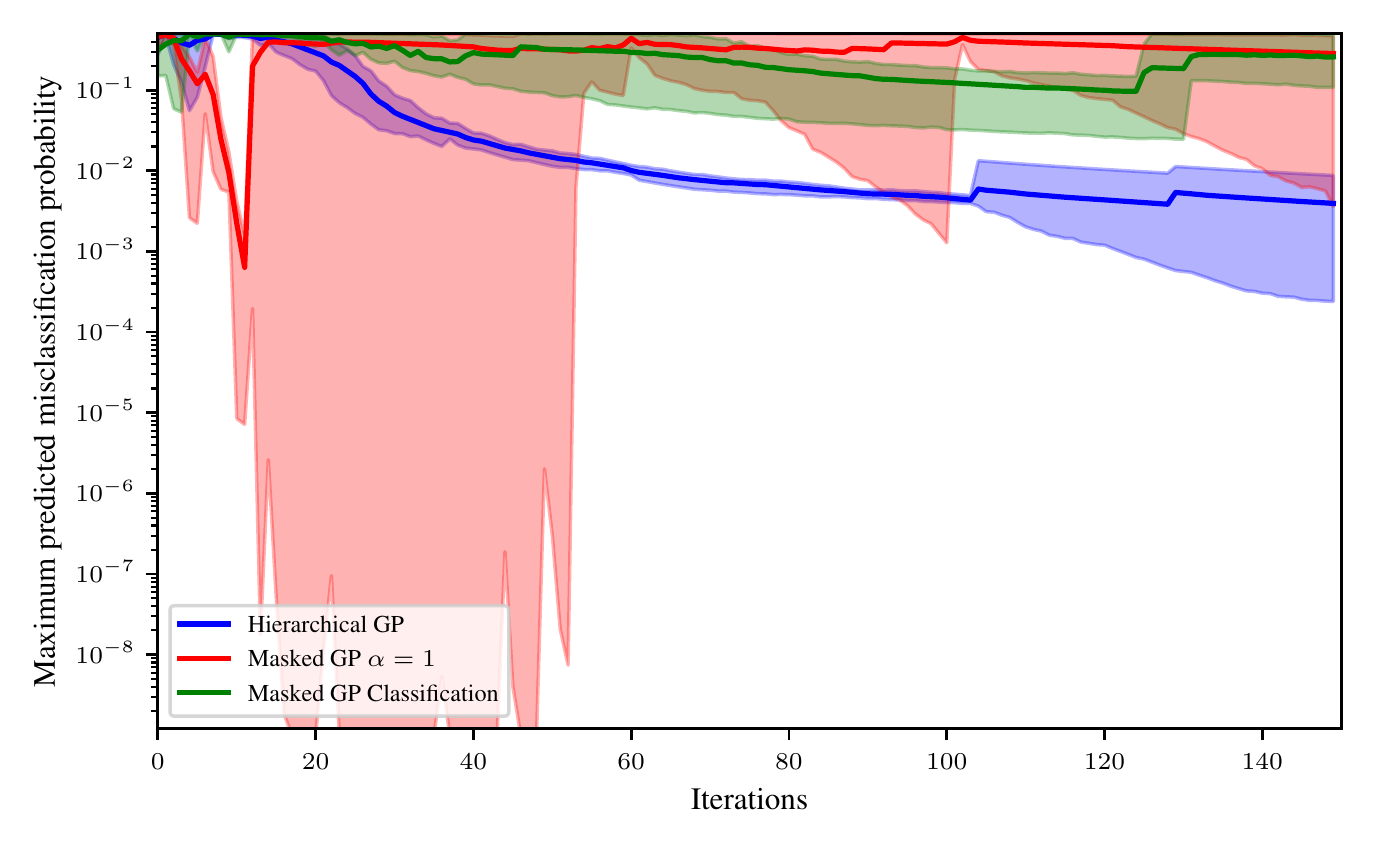}
   
     \end{subfigure}
        \caption{Convergence of failure probability ($p_f$), average precision, maximum predicted misclassification probability and \fscore score for fitted Gaussian processes for the simple toy function. The shaded area represents the minimum and maximum of 5 repeats, and the dark line represents the mean. All models eventually obtain a $p_f$ which is correct within the Monte Carlo error calculated using the CoV (approximately $0.0027$).
        }
        \label{fig:convergence}
\end{figure}

\begin{figure}
     \centering
     
     \begin{subfigure}[b]{0.49\textwidth}
         \centering
         \includegraphics[width=\textwidth]{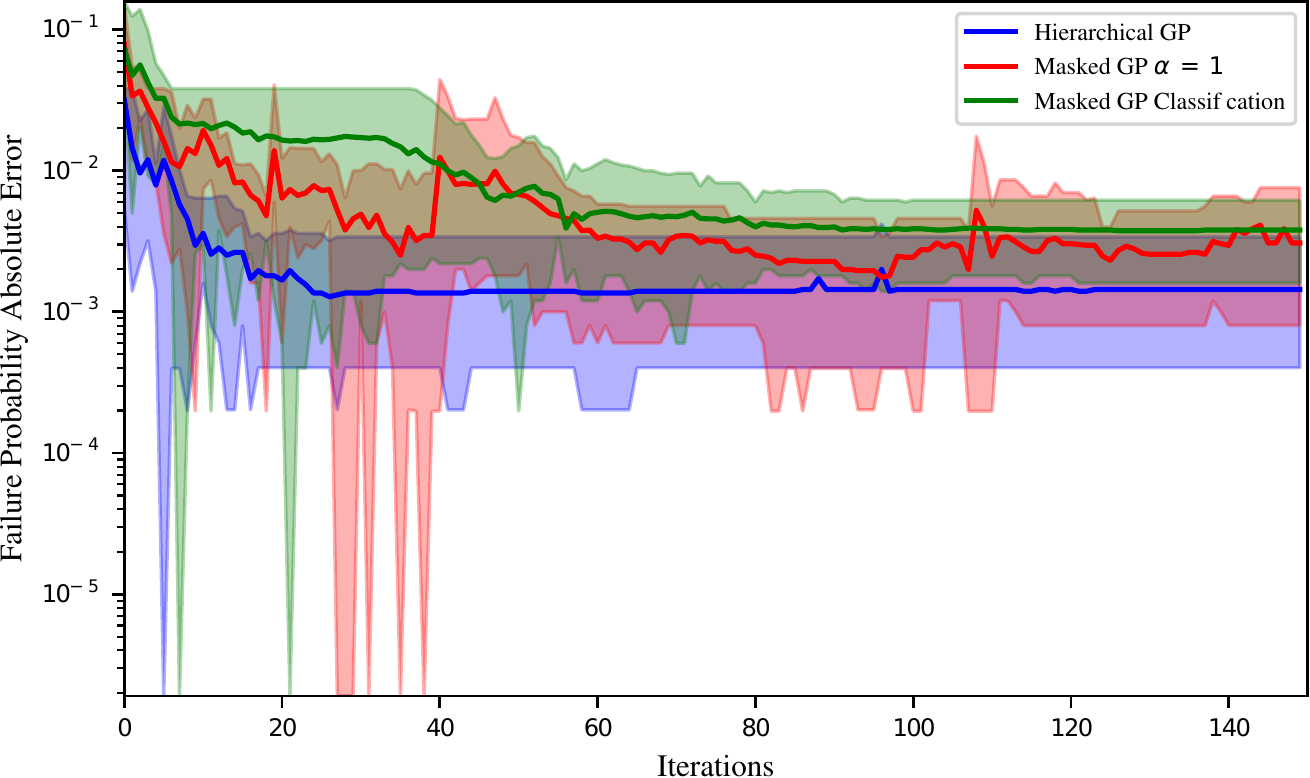}
     \end{subfigure}
\hfill
     \begin{subfigure}[b]{0.49\textwidth}
         \centering
         \includegraphics[width=\textwidth]{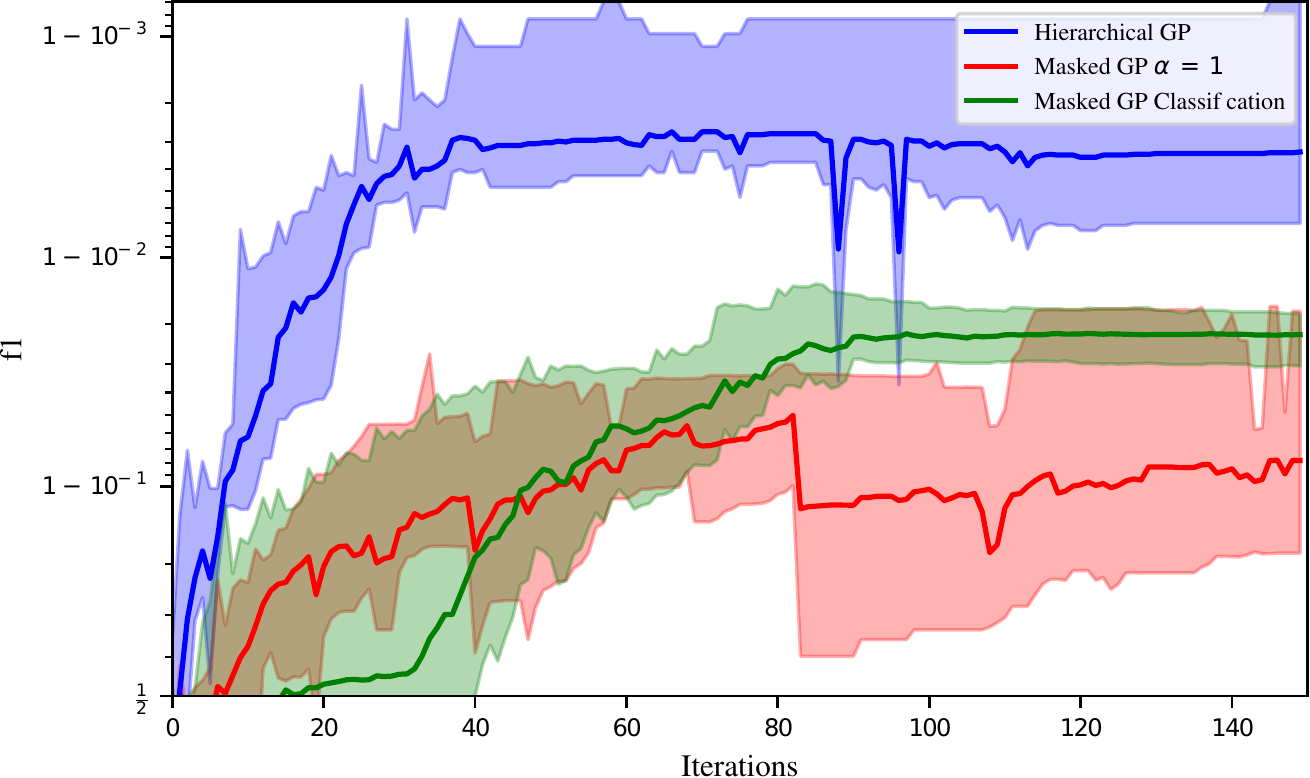}
     \end{subfigure}
     \\
     \begin{subfigure}[b]{0.49\textwidth}
         \centering
         \includegraphics[width=\textwidth]{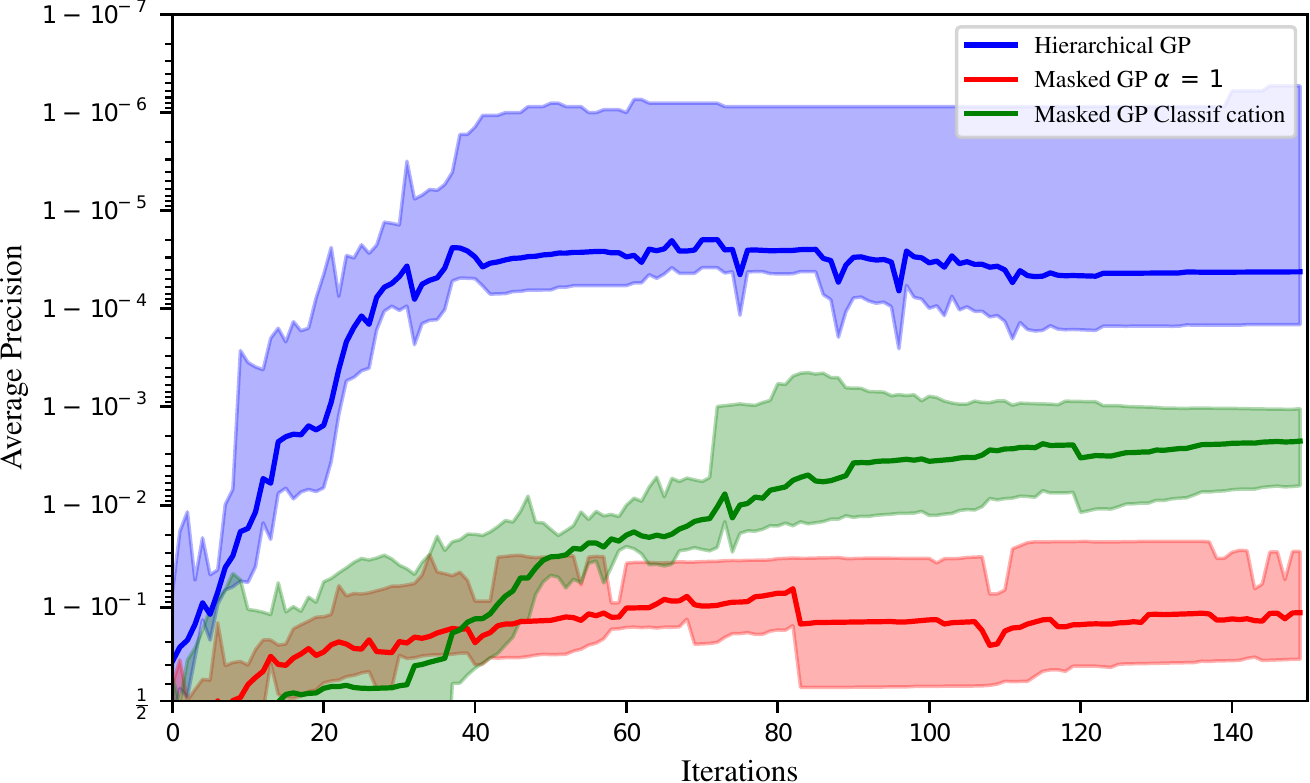}
     \end{subfigure}
     \hfill
     \begin{subfigure}[b]{0.49\textwidth}
         \centering
         \includegraphics[width=\textwidth]{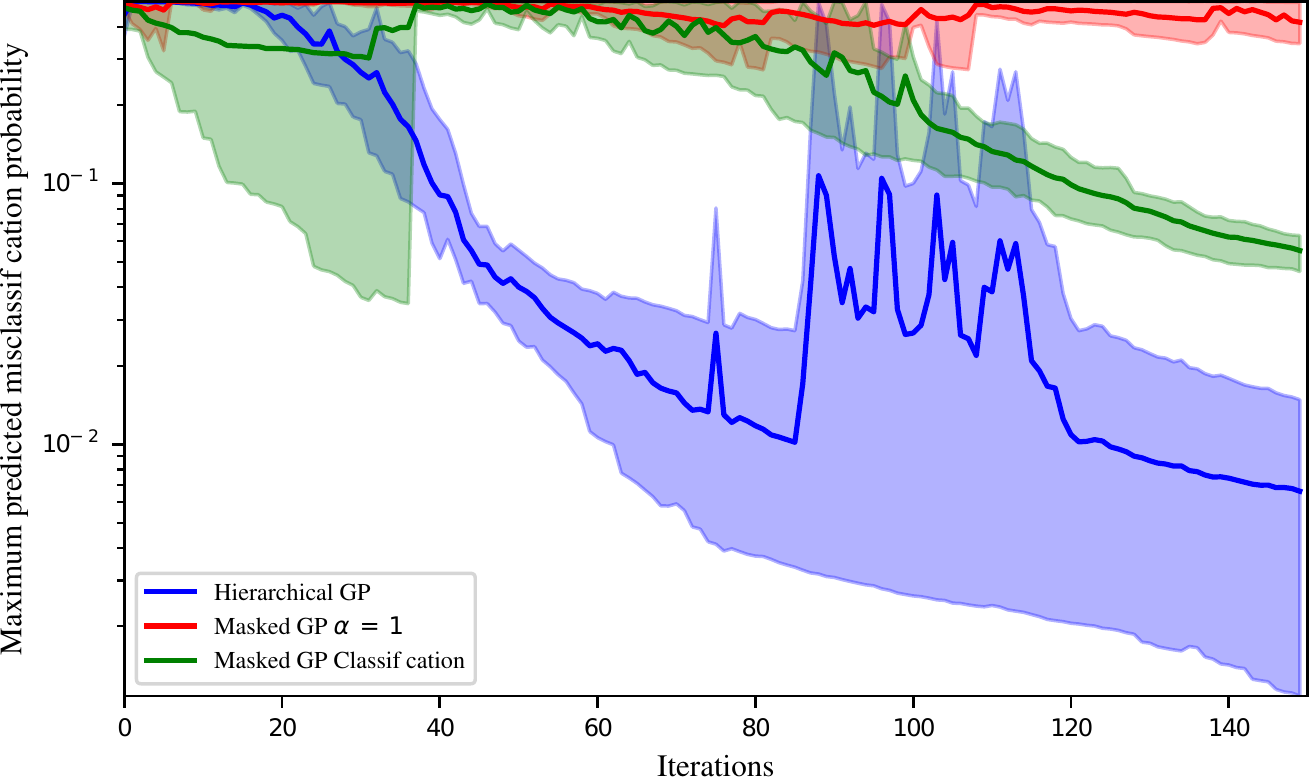}
     \end{subfigure}
     
             \caption{Convergence of failure probability ($p_f$), average precision, maximum predicted misclassification probability and \fscore score for fitted Gaussian processes for the autonomous driving experiment. The shaded area represents the minimum and maximum of 5 repeats, and the dark line represents the mean. All models eventually obtain a $p_f$ which is correct within the Monte Carlo error calculated using the CoV (approximately $0.0027$).
        }
        \label{fig:convergence2}
\end{figure}

Fig.~\ref{fig:convergence} and Fig.~\ref{fig:convergence2} show the convergence of the failure probability ($p_f$), maximum predicted misclassification probability, average precision and \fscore score for the models for the toy function and autonomous driving experiments.
In Table~\ref{tab:pf_terminate} we show the number of samples required for each algorithm to terminate and the $p_f$, average precision and \fscore score when the algorithm terminates.

\begin{table}
    \centering
    
    \caption{Comparison of $p_f$, coefficient of variation (CoV), \fscore score (\fscore), Average Precision (AP) and number of evaluations (N. Eval.) for all methodologies. The total function evaluations includes the initial design of experiments (12 evaluations) and the number of iterations of the active learning algorithm. Number of iterations was capped at 150 and methods hitting the cap are marked with did not terminate (DNT).
    Mean and standard deviation for 5 repeats are shown.
         }
		\begin{tabular}{@{}l@{\hspace{4pt}}l@{\hspace{4pt}}l@{\hspace{4pt}}l@{\hspace{4pt}}l@{\hspace{4pt}}l@{}}
		\toprule
         Methodology & $p_f$ & CoV & \fscore & AP & N. Eval. \\
         & Avg. (Std. Dev.) \\
         \midrule
         & \multicolumn{5}{c}{Toy function} \\
          \cmidrule{2-6}
         Analytic & 0.036906 & N/A & N/A & N/A & N/A \\ 
Hierarchical GP (ours) & 0.036 (0.0051) & 0.074 (0.0061) & 1 (0.0011) & 1 (6.1e-07) & 56 (3.4) \\
Masked GP $\alpha=1.0$ & 0.019 (0.0031) & 0.096 (0.013) & 0.66 (0.00013) & 0.56 (0.041) & 20 (3.2) \\
GPC & 0.037 (0.0026) & 0.073 (0.0027) & 0.99 (0.003) & 1 (0.00027) & DNT \\
          \cmidrule{2-6}
         & \multicolumn{5}{c}{Autonomous Driving} \\
          \cmidrule{2-6}
        Analytic & 0.0382 & N/A & N/A & N/A & N/A  \\
Hierarchical GP (ours) & 0.038 (0.0019) & 0.071 (0.0019) & 1 (0.0013) & 1 (1.7e-05) & 81 (21) \\
Masked GP $\alpha=1.0$ & 0.037 (0.0041) & 0.073 (0.0044) & 0.92 (0.063) & 0.89 (0.097) & DNT \\
GPC & 0.035 (0.0029) & 0.074 (0.003) & 0.98 (0.0049) & 1 (0.0023) & DNT \\
         \bottomrule
    \end{tabular}
    \label{tab:pf_terminate}
\end{table}

\subsection{Comparing values of $\alpha$ for the Regression GP}
\label{sec:alpha}

\begin{figure}
     \centering
     \begin{subfigure}[b]{0.49\textwidth}
         \centering
         \includegraphics[width=\textwidth]{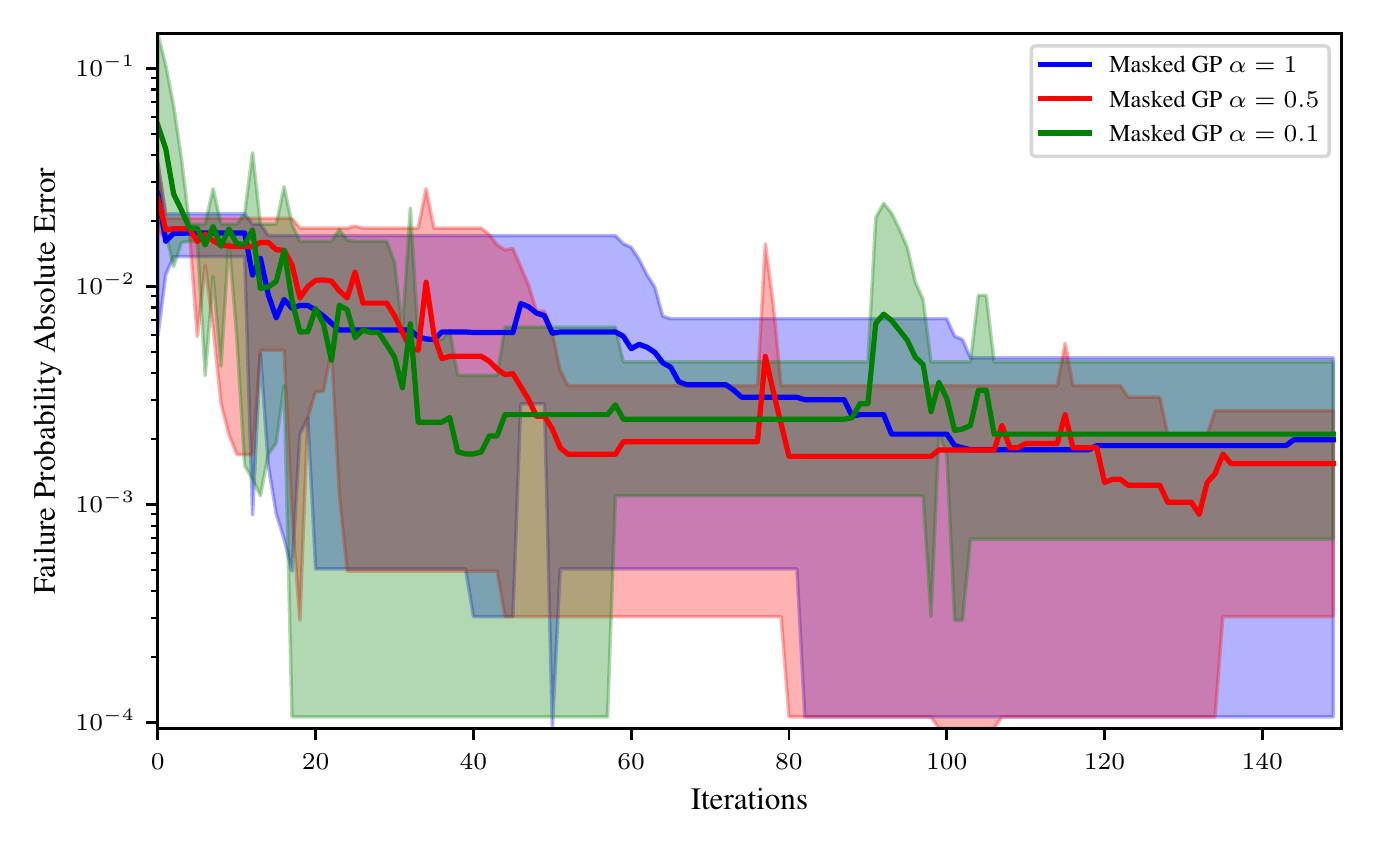}
     \end{subfigure}
     \hfill
     \begin{subfigure}[b]{0.49\textwidth}
         \centering
         \includegraphics[width=\textwidth]{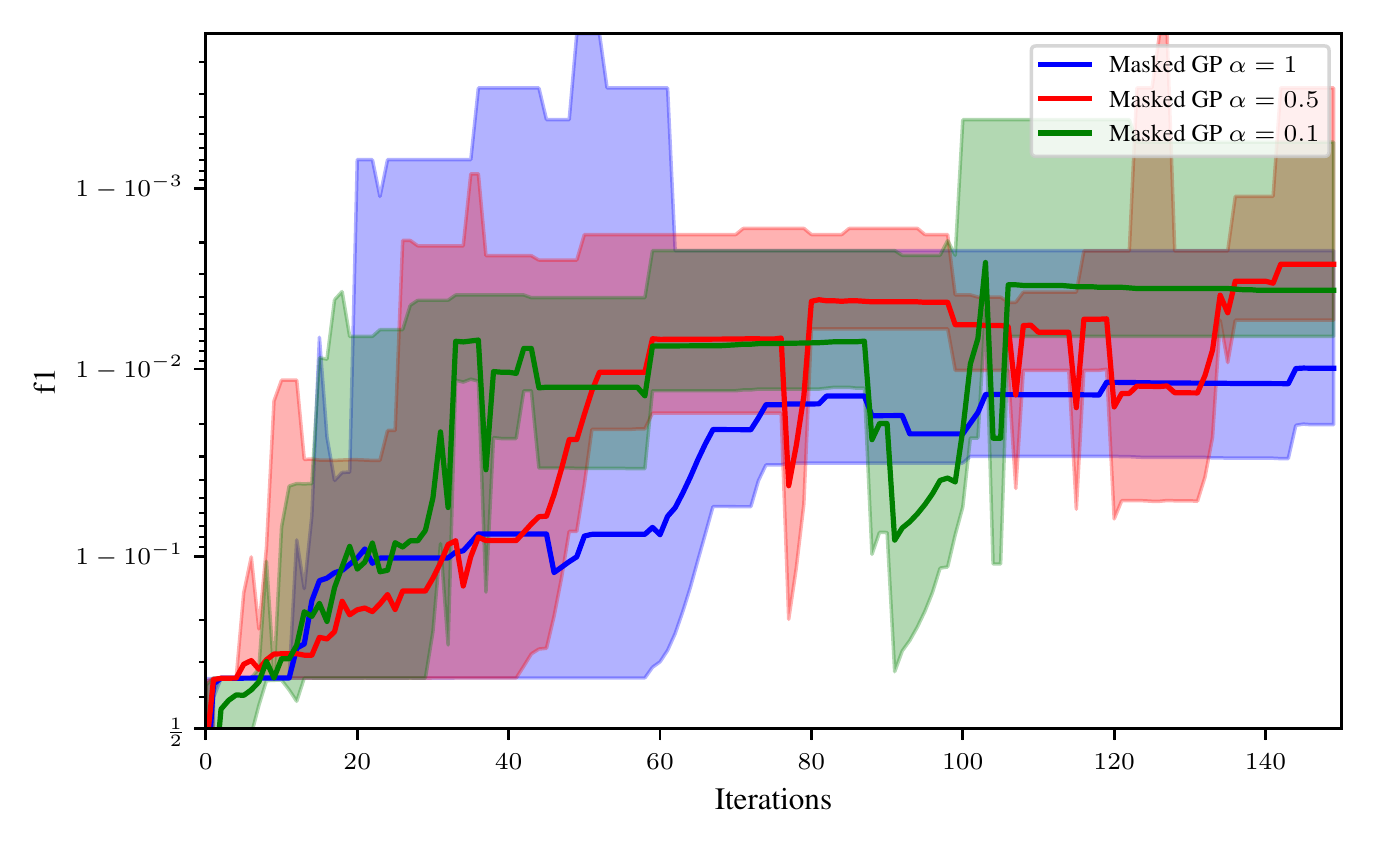}
     \end{subfigure}\\
          \begin{subfigure}[b]{0.49\textwidth}
         \centering
         \includegraphics[width=\textwidth]{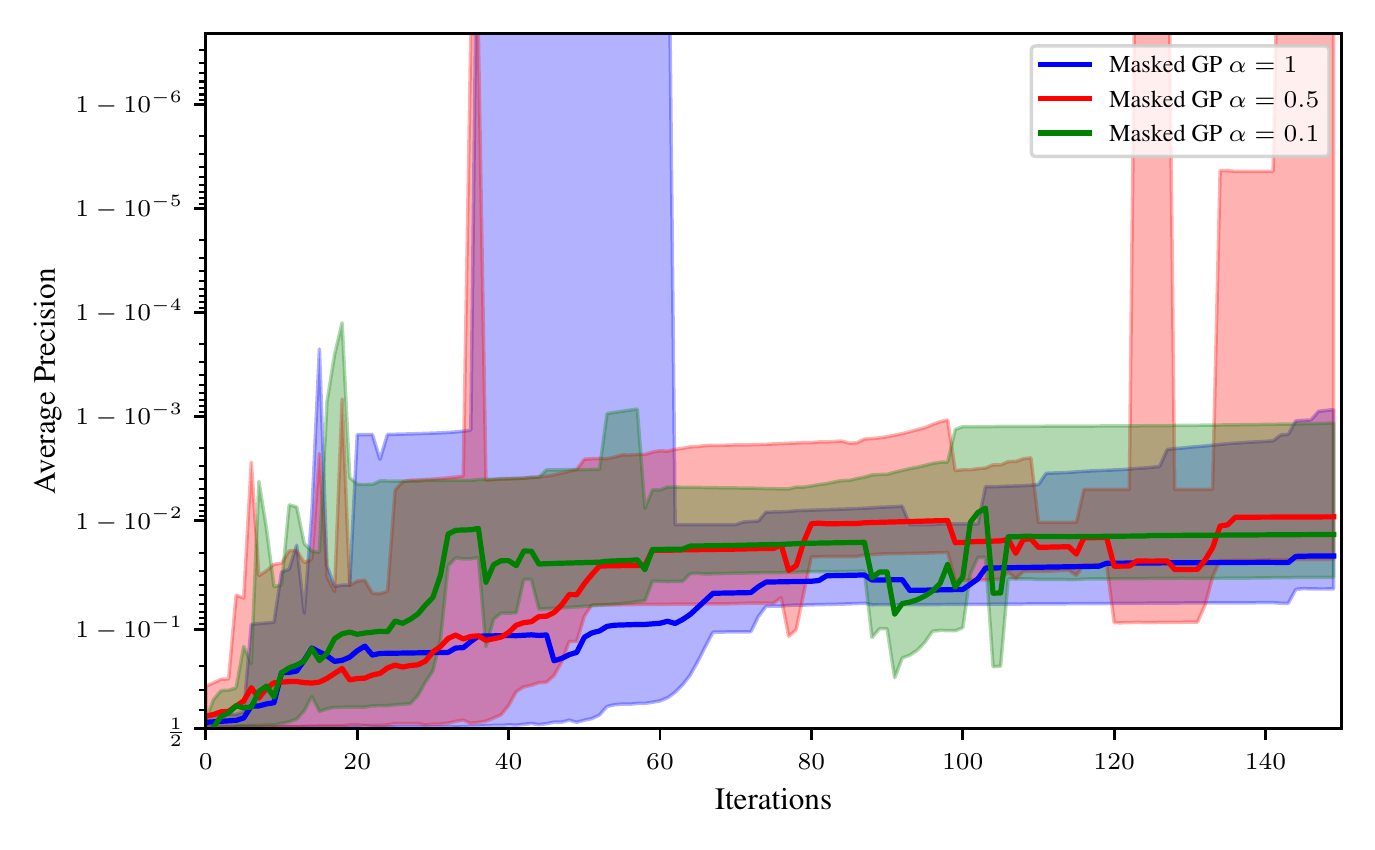}
   
     \end{subfigure}
          \hfill
     \begin{subfigure}[b]{0.49\textwidth}
         \centering
         \includegraphics[width=\textwidth]{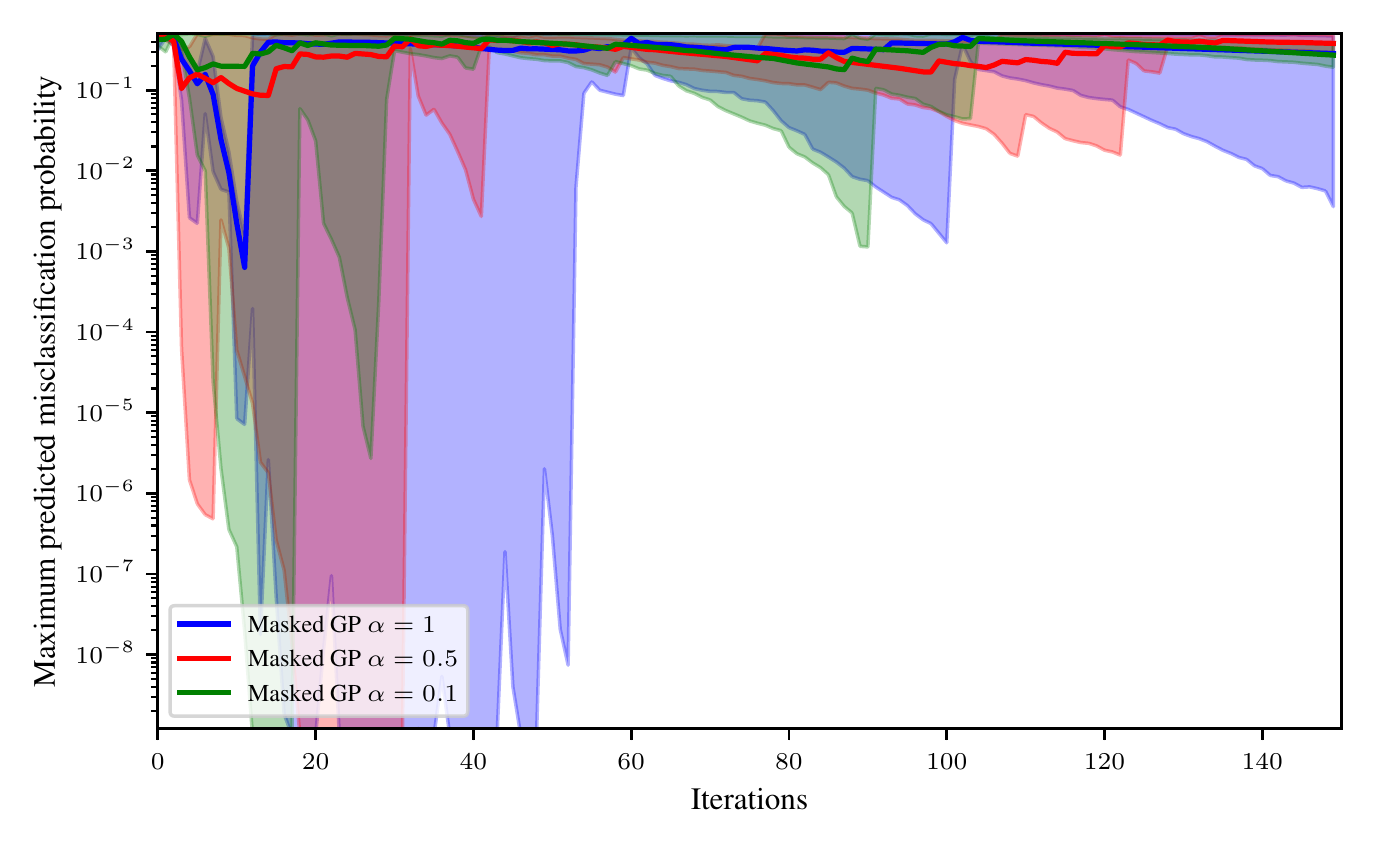}
   
     \end{subfigure}
        \caption{Convergence of failure probability ($p_f$), average precision, maximum predicted misclassification probability and \fscore score for fitted masked Gaussian processes with different $\alpha$ for the simple toy function. The shaded area represents the minimum and maximum of 5 repeats, and the dark line represents the mean. All models eventually obtain a $p_f$ which is correct within the Monte Carlo error calculated using the CoV (approximately $0.0027$).
        }
        \label{fig:convergence_alpha}
\end{figure}

\begin{figure}
     \centering
     
     \begin{subfigure}[b]{0.49\textwidth}
         \centering
         \includegraphics[width=\textwidth]{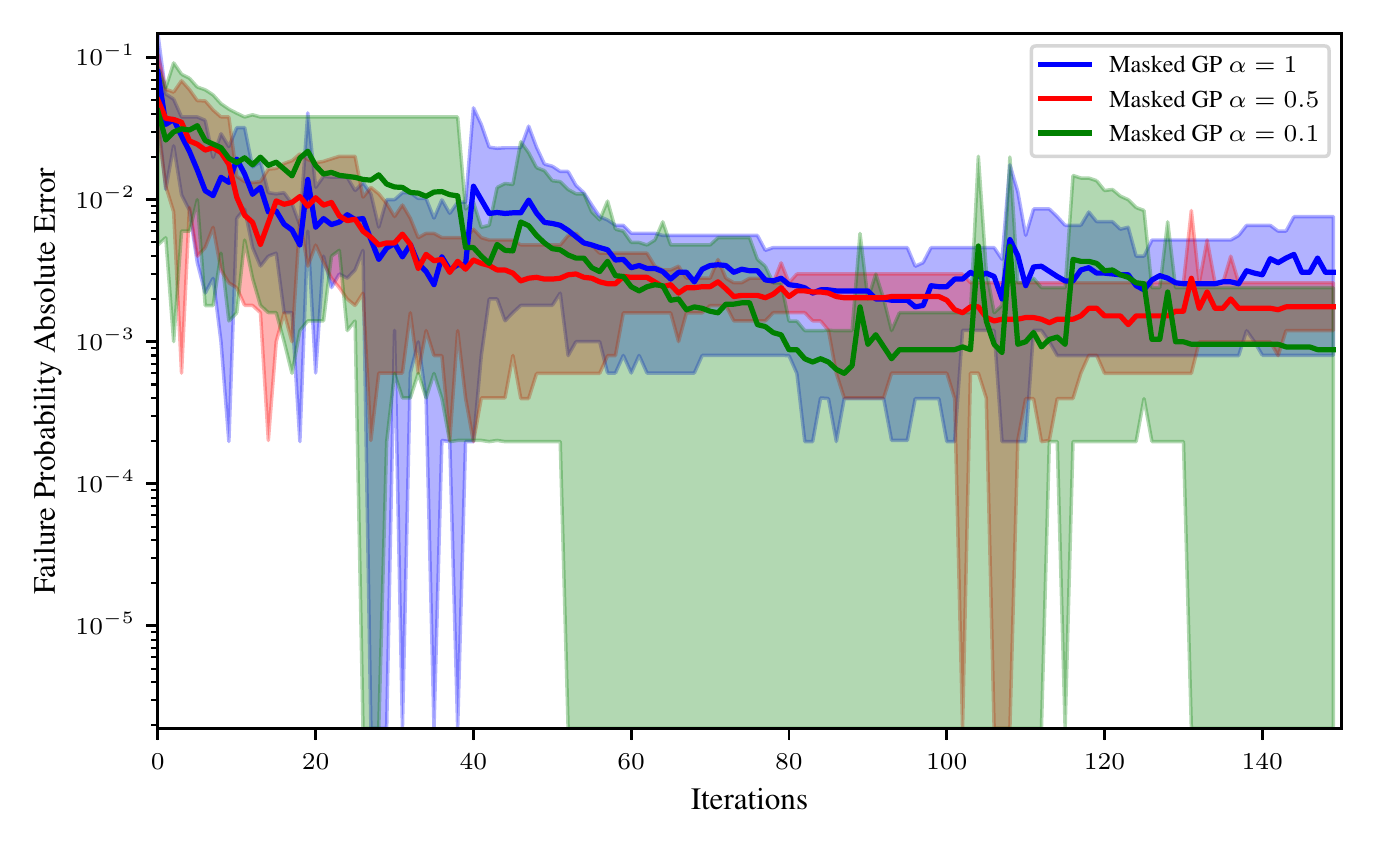}
     \end{subfigure}
\hfill
     \begin{subfigure}[b]{0.49\textwidth}
         \centering
         \includegraphics[width=\textwidth]{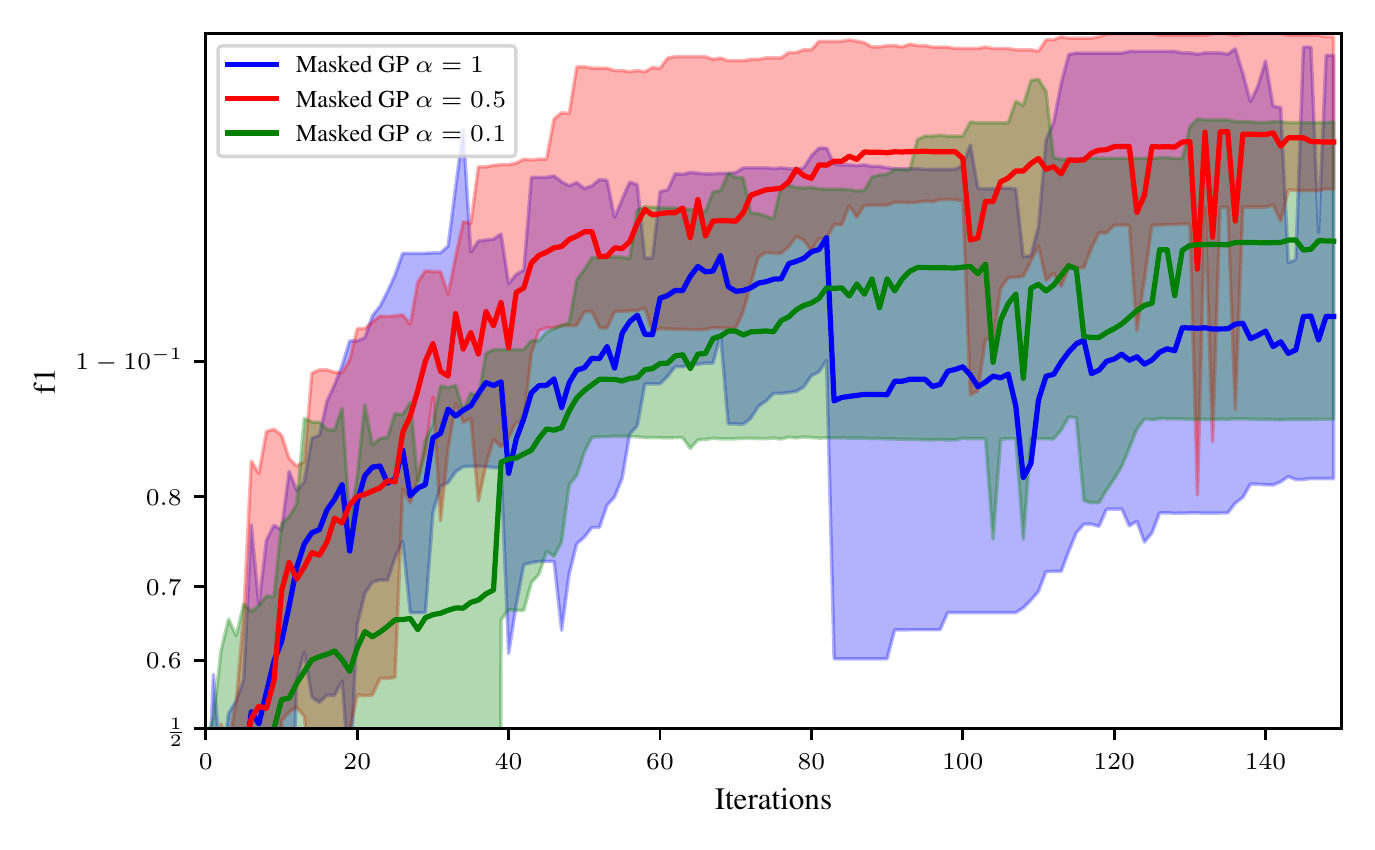}
     \end{subfigure}
     \\
     \begin{subfigure}[b]{0.49\textwidth}
         \centering
         \includegraphics[width=\textwidth]{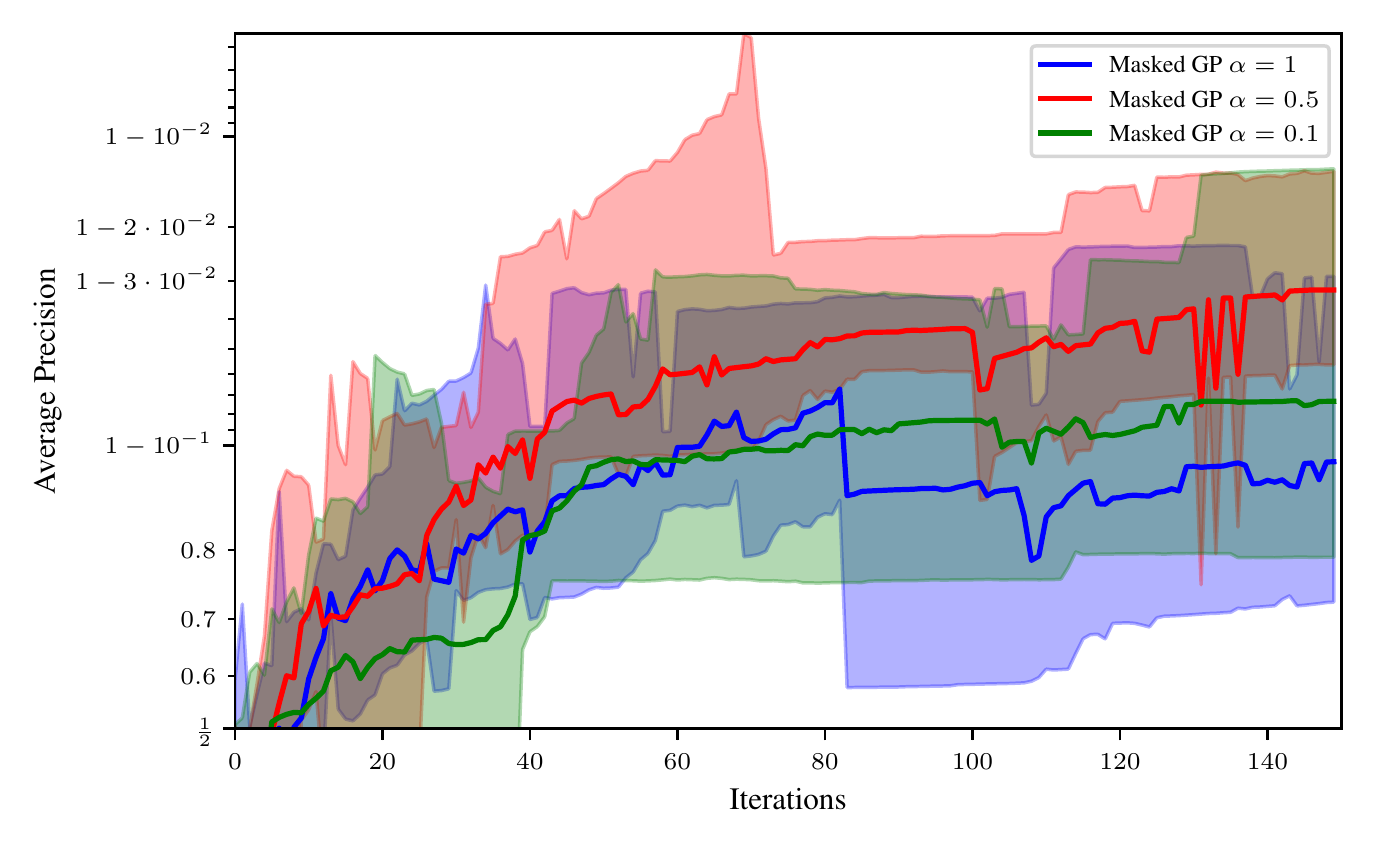}
     \end{subfigure}
     \hfill
     \begin{subfigure}[b]{0.49\textwidth}
         \centering
         \includegraphics[width=\textwidth]{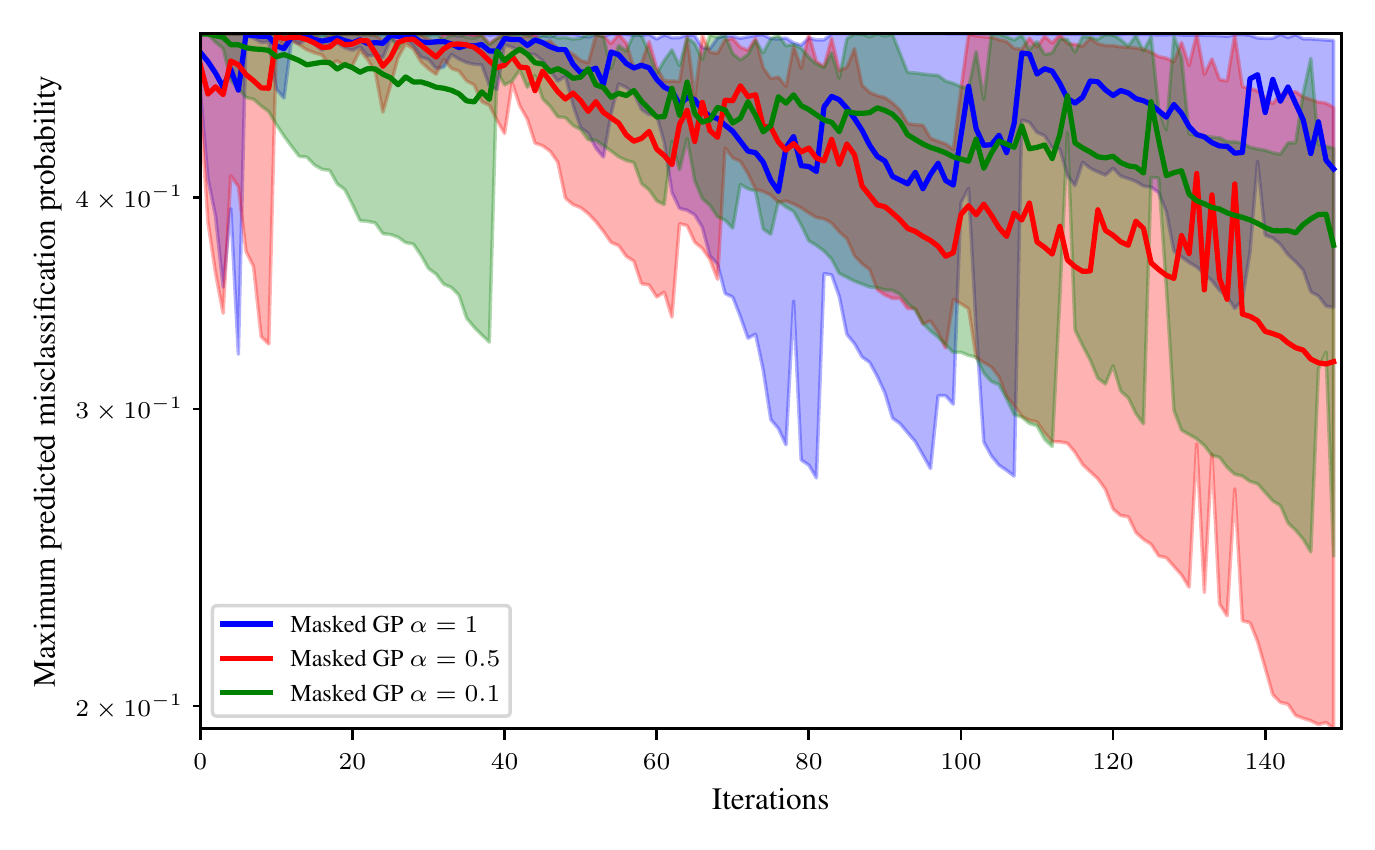}
     \end{subfigure}
     
             \caption{Convergence of failure probability ($p_f$), average precision, maximum predicted misclassification probability and \fscore score for fitted masked Gaussian processes with different $\alpha$ for the autonomous driving experiment. The shaded area represents the minimum and maximum of 5 repeats, and the dark line represents the mean. All models eventually obtain a $p_f$ which is correct within the Monte Carlo error calculated using the CoV (approximately $0.0027$).
        }
        \label{fig:convergence2_alpha}
\end{figure}

In Fig.~\ref{fig:convergence_alpha}-\ref{fig:convergence2_alpha} we repeat the analysis from Section~\ref{sec:experiments} to show the convergence of failure probability ($p_f$), average precision, maximum predicted misclassification probability and \fscore score for fitted masked Gaussian processes with different $\alpha$.
In comparison to Hierarchical GP, all masked GPs are similar.
All models show slower convergence than Hierarchical GP.
Table~\ref{tab:pf_terminate_alpha} shows the termination results for different values of $\alpha$.

\begin{table}
    \centering
    
    \caption{Comparison of $p_f$, coefficient of variation (CoV), \fscore score (\fscore), Average Precision (AP) and number of evaluations (N. Eval.) for fitted masked Gaussian processes with different $\alpha$. The total function evaluations includes the initial design of experiments (12 evaluations) and the number of iterations of the active learning algorithm. Number of iterations was capped at 150 and methods hitting the cap are marked with did not terminate (DNT). 
    Mean and standard deviation for 5 repeats are shown.
         }
		\begin{tabular}{@{}l@{\hspace{4pt}}l@{\hspace{4pt}}l@{\hspace{4pt}}l@{\hspace{4pt}}l@{\hspace{4pt}}l@{}}
		\toprule
         Methodology & $p_f$ & CoV & \fscore & AP & N. Eval. \\
         & Avg. (Std. Dev.) \\
         \midrule
         & \multicolumn{5}{c}{Toy function} \\
          \cmidrule{2-6}
Masked GP $\alpha=0.1$ & 0.025 (0.0089) & 0.08 (0.0098) & 0.79 (0.19) & 0.71 (0.26) & 75 (78) \\
Masked GP $\alpha=0.5$ & 0.018 (0.0012) & 0.086 (0.018) & 0.66 (6.7e-05) & 0.56 (0.055) & 19 (4.9) \\
Masked GP $\alpha=1.0$ & 0.019 (0.0031) & 0.096 (0.013) & 0.66 (0.00013) & 0.56 (0.041) & 20 (3.2) \\
          \cmidrule{2-6}
         & \multicolumn{5}{c}{Autonomous Driving} \\
          \cmidrule{2-6}
Masked GP $\alpha=0.1$ & 0.038 (0.0014) & 0.071 (0.0014) & 0.95 (0.047) & 0.93 (0.078) & DNT \\
Masked GP $\alpha=0.5$ & 0.038 (0.0019) & 0.072 (0.0019) & 0.97 (0.0093) & 0.97 (0.017) & DNT \\
Masked GP $\alpha=1.0$ & 0.037 (0.0041) & 0.073 (0.0044) & 0.92 (0.063) & 0.89 (0.097) & DNT \\
         \bottomrule
    \end{tabular}
    \label{tab:pf_terminate_alpha}
\end{table}

\end{document}